\def\year{2022}\relax
\documentclass[letterpaper]{article} 
\usepackage{aaai22}  
\usepackage{times}  
\usepackage{helvet}  
\usepackage{courier}  
\usepackage[hyphens]{url}  
\usepackage{graphicx} 
\usepackage{natbib}  
\usepackage{caption} 
\DeclareCaptionStyle{ruled}{labelfont=normalfont,labelsep=colon,strut=off} 
\usepackage[utf8]{inputenc} 
\usepackage[T1]{fontenc}    
\usepackage{booktabs}       
\usepackage{amsfonts}       
\usepackage{nicefrac}       
\usepackage{microtype}      
\usepackage{xspace}
\usepackage{subfig}
\usepackage{graphicx}

\usepackage{amsmath,amsthm,amsfonts,amssymb,amscd}
\usepackage{cleveref}
\usepackage{natbib}
\usepackage{url}
\usepackage{paralist}
\usepackage{enumitem}
\usepackage{enumerate}
\usepackage{algorithm}
\usepackage{multirow}
\usepackage{xcolor}
\usepackage{placeins}

\newcommand{\cifar}{{CIFAR-10}\xspace}
\newcommand{\imagenet}{{ImageNet-1K}\xspace}

\newcommand{\sync}{{SSGD}\xspace}
\newcommand{\ssgd}{{SSGD}\xspace}

\newcommand{\dpsgd}{{DPSGD}\xspace}
\newcommand{\eff}{{EfficientNet-B0}\xspace}
\newcommand{\vgg}{{VGG-19}\xspace}
\newcommand{\resnet}{{ResNet-18}\xspace}
\newcommand{\densenet}{{DenseNet-121}\xspace}
\newcommand{\googlenet}{{GoogleNet}\xspace}
\newcommand{\mobilenet}{{MobileNet}\xspace}
\newcommand{\mobilenetvtwo}{{MobileNetV2}\xspace}
\newcommand{\shuf}{{ShuffleNet}\xspace}
\newcommand{\resnext}{{ResNext-29}\xspace}
\newcommand{\senet}{{SENet-18}\xspace}
\newcommand{\swba}{{SWB-300}\xspace}
\newcommand{\swbb}{{SWB-2000}\xspace}
\newcommand{\weiz}[1]{\textcolor{black}{#1}}

\newcommand{\mingrui}[1]{\textcolor{black}{#1}}

\usepackage{aaai22}

\title{Loss Landscape Dependent Self-Adjusting Learning Rates in Decentralized Stochastic Gradient Descent}
\author{%
 Wei Zhang$^{1}$ Mingrui Liu$^{2}$ Yu Feng$^{3}$ Xiaodong Cui$^{1}$ Brian Kingsbury$^{1}$ Yuhai Tu$^{1}$\\
 \texttt{weiz@us.ibm.com mingruiliu.ml@gmail.com yufeng.physics@gmail.com cuix@us.ibm.com bedk@us.ibm.com yuhai@us.ibm.com} \\
 IBM Research$^{1}$ George Mason University$^{2}$ Duke Unversity$^{3}$\\
}
\begin{document}
\maketitle
\begin{abstract}
Distributed Deep Learning (DDL) is essential for large-scale Deep Learning (DL) training. \weiz{Synchronous Stochastic Gradient Descent (SSGD)~\footnote{In the literature, SSGD is also called "Centralized Synchronized Stochastic Gradient Descent". In this paper, we use these two terms interchangeably.} is the de facto DDL optimization method. Using a sufficiently large batch size is critical to achieving DDL runtime speedup. In a large batch setting, the learning rate must be increased to compensate for the reduced number of parameter updates. However, a large learning rate may harm convergence in SSGD and training could easily diverge.}  \weiz{Recently, Decentralized Parallel SGD (DPSGD) has been proposed to improve distributed training speed. 
In this paper, we find that DPSGD not only has a system-wise runtime benefit but also a significant convergence benefit over SSGD in the large batch setting.}
Based on a detailed analysis of the DPSGD learning dynamics, we find that DPSGD introduces additional landscape-dependent noise that automatically adjusts the effective learning rate to improve convergence. In addition, we theoretically show that this noise smoothes the loss landscape, hence allowing a larger learning rate. We conduct extensive studies over 18 state-of-the-art DL models/tasks and demonstrate that DPSGD often converges in cases where SSGD diverges for large learning rates in the large batch setting. Our findings are consistent across two different application domains: Computer Vision (CIFAR10 and ImageNet-1K) and Automatic Speech Recognition (SWB300 and SWB2000), and two different types of neural network models: Convolutional Neural Networks and Long Short-Term Memory Recurrent Neural Networks.
\end{abstract}
\section{Introduction}
\label{sec:intro}
Deep Learning (DL) has revolutionized AI training across application domains: Computer Vision (CV) \cite{alexnet, resnet}, Natural Language Processing (NLP) \citep{transformer}, and Automatic Speech Recognition (ASR) \citep{asr-dl}. Stochastic Gradient Descent (SGD) is the fundamental optimization method used in DL training. Due to massive computational requirements, Distributed Deep Learning (DDL) is the preferred mechanism to train large scale Deep Learning (DL) tasks.

The degree of parallelism in a DDL system is dictated by batch size: the larger the batch size, the more parallelism and higher speedup can be expected. However, large batches require a larger learning rate and overall they may negatively affect model accuracy because
\begin{inparaenum}[(1)]
\item large batch training usually converges to sharp minima which do not generalize well~\cite{keskar2016large}, and 
\item large learning rates may violate the conditions (i.e., the learning rate should be less than the reciprocal of the smoothness parameter) required for convergence in nonconvex optimization theory~\citep{ghadimi2013stochastic}.
\end{inparaenum}
Although training longer with large batches can lead to better generalization~\citep{hoffer2017train}, doing so gives up some or all of the speedup we seek. Through meticulous hyper-parameter design (e.g., learning rate schedules) tailored to each specific task, SSGD-based DDL systems have enabled large batch training and shortened training time for some challenging CV tasks~\citep{facebook-1hr, lars} and NLP tasks \citep{lamb} from weeks to hours or less.  However, it is observed that SSGD with large batch size leads to large training loss and inferior model quality for ASR tasks~\citep{interspeech19}, as illustrated in \Cref{fig:swb300-noise} (red curve). Here, we found for other types of tasks (e.g. CV) and DL models, large batch SSGD has the same problem (\Cref{fig:eff-noise} and \Cref{fig:senet-noise}). 

Several SSGD variants have been proposed to address large batch training problems:
\begin{inparaenum}[(1)]
\item local SGD, i.e., SGD-based algorithms with periodic averaging, where learners conduct global averaging after multiple steps of gradient-based updates~\cite{swap,localsgd,kstep};
\item SSGD based algorithm with second-order statistics, including adaptive gradient algorithms~\cite{lamb, lars} and algorithms for exploring the information from the gradient covariance matrix~\cite{covnoise}; and
\item SSGD-based algorithms on a smoothed landscape~\cite{extragrad,sam}, in which specifically designed loss landscape smoothing algorithms are used.
\end{inparaenum}
All of these approaches require global synchronization and/or global statistics collection, which makes them vulnerable to stragglers. 

Decentralized algorithms, such as Decentralized Parallel Stochastic Gradient Descent (DPSGD)~\cite{dpsgd}, are surrogates for SSGD in machine learning. Unlike SSGD, where each learner updates its weights by taking a global average of all learners' weights, DPSGD updates each learner's weights by taking a partial average (i.e., across a subset of neighboring learners). In contrast to the existing variants of SSGD, DPSGD requires no additional calculation and no global synchronization. Traditionally DPSGD is a second-choice to SSGD, and is used only when the underlying computational resources are less homogeneous (i.e., a high latency network or computational devices running at different speeds). Little thought has been given to the question of whether there are any convergence benefits for DPSGD,  especially in the large batch setting.


In this paper, we find that DPSGD~\citep{lian2017can} greatly improves large batch training performance, as illustrated by the green curves in \Cref{fig:noise}. Since DPSGD only uses a partial average of neighboring learners' weights, each learner's weights differ from the weights of other learners. The differing weights between learners are an additional source of noise in DPSGD training.
The key difference between SSGD, SSGD with Gaussian noise (denoted as "SSGD$^*$" in this paper) 
and DPSGD is the source of noise during the update, and this noise directly affects performance in deep learning. This naturally motivates us to ask \textit{Why does decentralized training outperform synchronous training in the large batch setting?} More specifically, we try to understand whether these performance differences are caused by differences in noise. 
We answer this question from both theoretical and empirical perspectives. Our contributions are:

\begin{figure}[t]
\small
    \centering
    \subfloat[\parbox{0.32\linewidth}{\tiny LSTM, SWB300, BS8192}]{
      {\includegraphics[width=0.33\columnwidth]{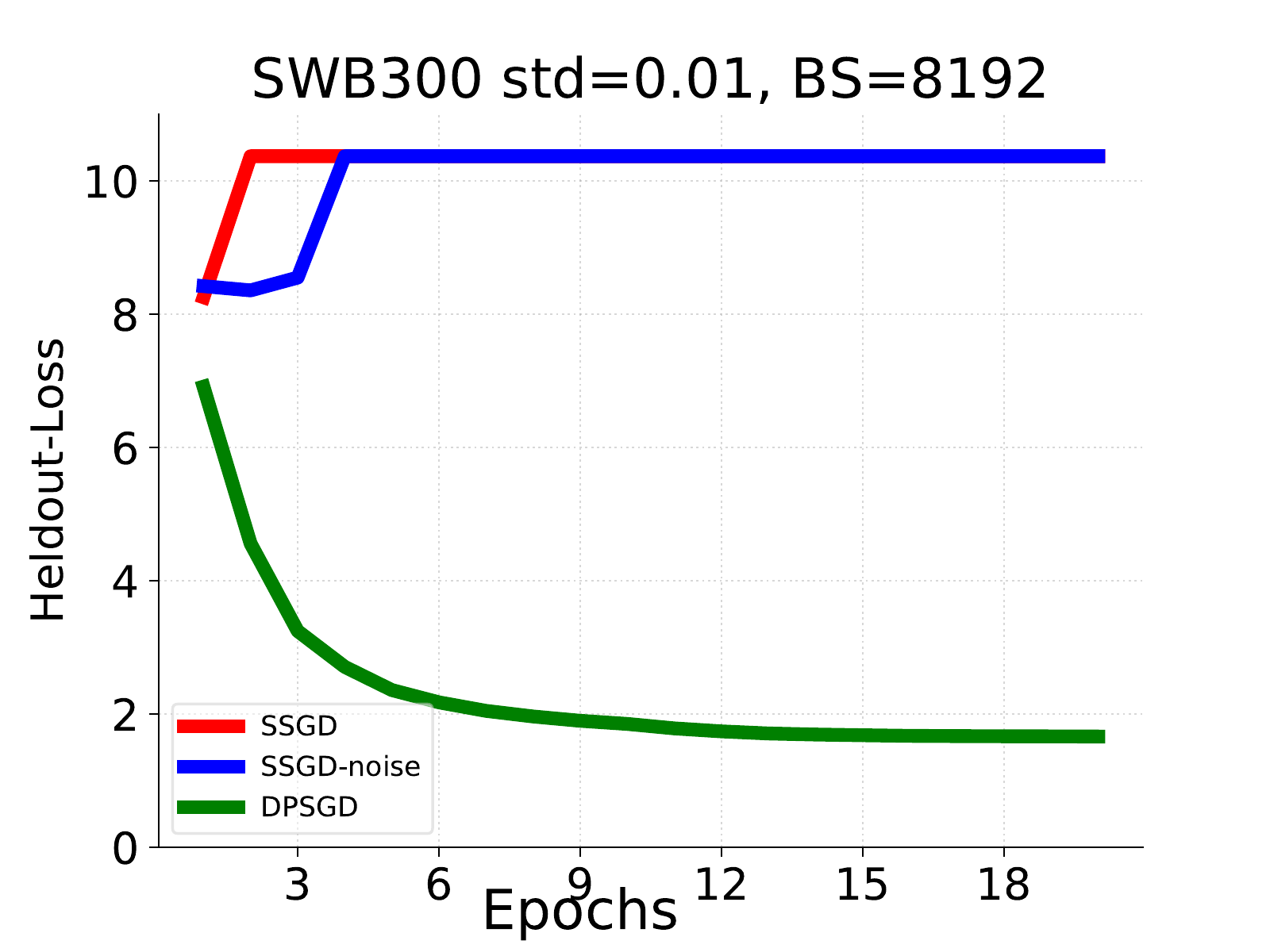}
      }\label{fig:swb300-noise}
    }
    \subfloat[\parbox{0.32\linewidth}{\tiny EfficientNet, CIFAR-10, BS8192}]{
      {\includegraphics[width=0.33\columnwidth]{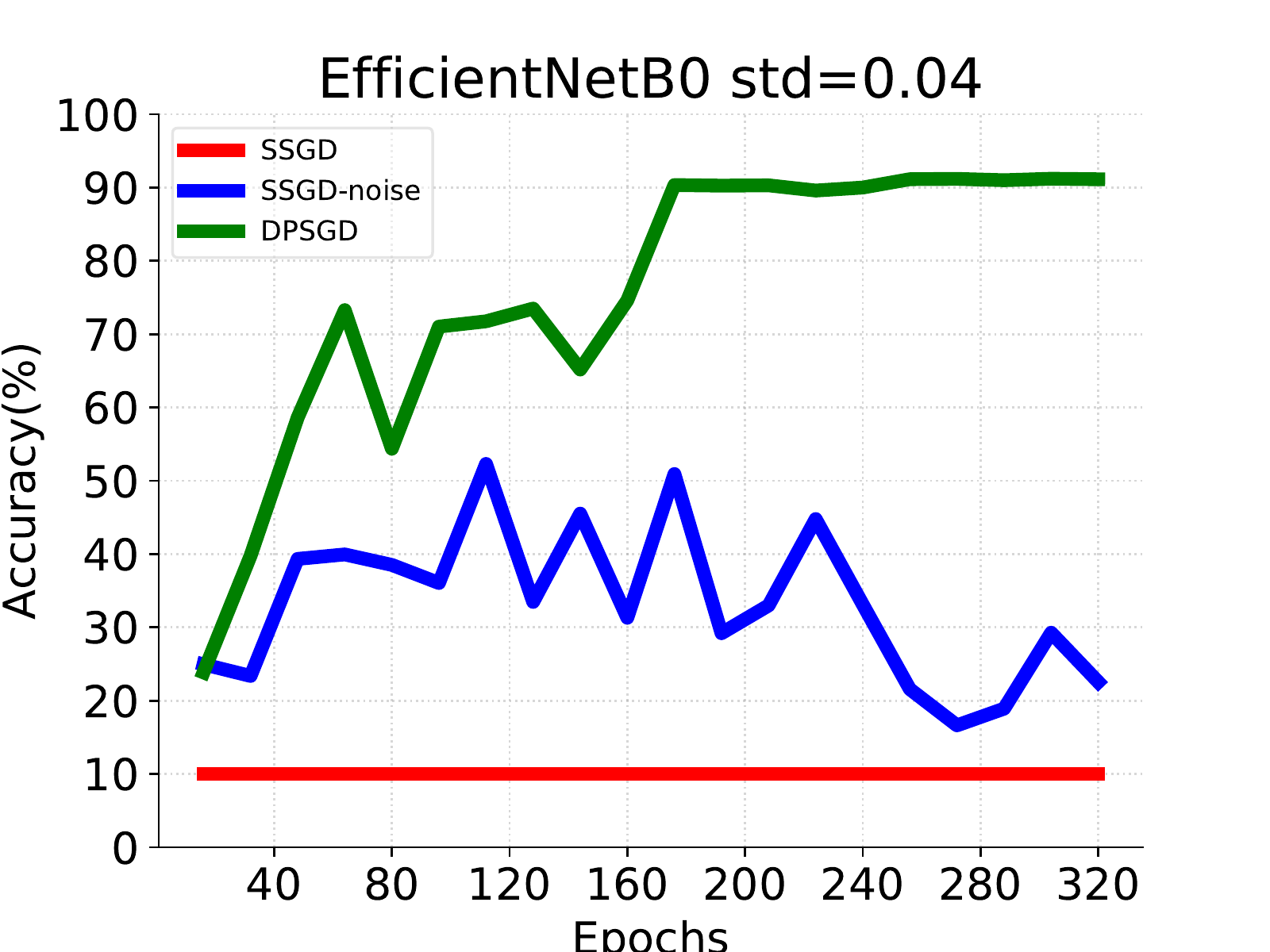}
      }\label{fig:eff-noise}
    } 
    \subfloat[\parbox{0.32\linewidth}{\tiny \senet, CIFAR-10, BS8192}]{
      {\includegraphics[width=0.33\columnwidth]{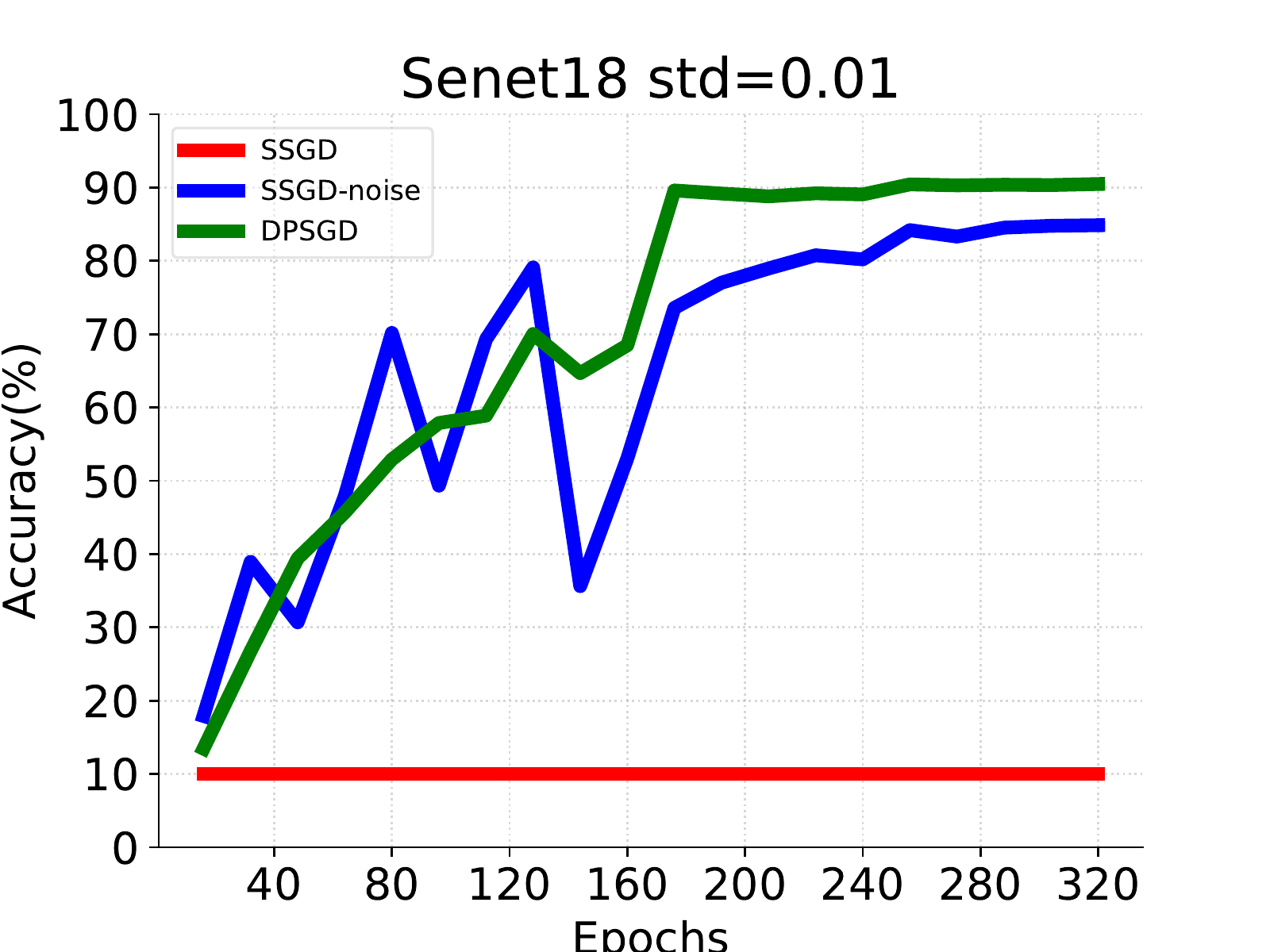}
      }\label{fig:senet-noise}
    }
    
\caption{SSGD (red) does not converge in the large batch setting.  \Cref{fig:swb300-noise} plots the heldout-loss, the lower the better. \Cref{fig:eff-noise} and \Cref{fig:senet-noise} plot the model accuracy, the higher the better. By injecting Gaussian noise, \sync might escape early traps but results in a much worse model (blue) compared to \dpsgd (green) in the large batch setting. The detailed task descriptions and training recipes are described in \Cref{sec:noise-lr-tuning}. BS stands for Batch-Size.}
    \label{fig:noise}
\end{figure}

\begin{itemize}[noitemsep,topsep=0pt,parsep=0pt,partopsep=0pt]
\item We analyze the dynamics of DDL algorithms, including both SSGD and DPSGD. We show, both theoretically and empirically, that the \textit{intrinsic noise} in DPSGD automatically adjusts the effective learning rate when the batch size is large to help convergence. Note that the intrinsic noise comes completely for free in the DPSGD algorithm, and we show that it has a loss-landscape smoothing effect. 
\item We conduct extensive empirical studies of 18 CV and ASR tasks with state-of-the-art CNN and LSTM models. Our experimental results demonstrate that \dpsgd consistently outperforms \sync, across application domains and Neural Network (NN) architectures in the large batch setting, \textit{without any hyper-parameter tuning}. To the best of our knowledge, DPSGD is the only generic algorithm that can improve SSGD large batch training on this many models/tasks. Furthermore, DPSGD does not require any global synchronization, unlike other solutions.

\end{itemize}

The remainder of this paper is organized as follows. \Cref{sec:theory} details the problem formulation and learning dynamics analysis of SSGD, SSGD$^*$, and DPSGD; \Cref{sec:meth} and \Cref{sec:results} detail the empirical results; \Cref{sec:related} discusses related work; and \Cref{sec:conclusion} concludes the paper.
  
 


\newcommand {\be}{\begin{equation}}
\newcommand {\ee}{\end{equation}}
\newcommand {\bea}{\begin{eqnarray}}
\newcommand {\eea}{\end{eqnarray}}
\newcommand{\x}{\mathbf{x}}
\newcommand{\g}{\mathbf{g}}
\newtheorem{thm}{Theorem}
\section{Analysis of stochastic learning dynamics in SSGD and DPSGD}
\label{sec:theory}

We first formulate the dynamics of an SGD based learning algorithm with multiple ($n >1$) learners indexed by $j=1,2,3,...n$ following the same theoretical framework established for a single learner~\citep{Chaudhari_2018}. At time (iteration) $t$, each learner has its own weight vector $\vec{w}_j(t)$, and the average weight vector $\vec{w}_a(t)$ is defined as: 
$\vec{w}_a(t)\equiv n^{-1}\sum_{j=1}^{n} \vec{w}_j(t)$.
Each learner $j$ updates its weight vector according to the cross-entropy loss function $L^{\mu_j(t)}(\vec{w})$ for minibatch $\mu_j(t)$ that is assigned to it at time $t$. The size of the local minibatch is $B$, and the overall batch size for all learners is $nB$. Two multi-learner algorithms, SSGD and DPSGD, are described below. 

{\bf (1) Synchronous Stochastic Gradient Descent (SSGD):} In the synchronous algorithm, each learner $j\in[1,n]$ starts from the average weight vector $\vec{w}_a$ and moves along the gradient of its local loss function  $L^{\mu_j(t)}$ evaluated at the average weight $\vec{w}_a$:
\begin{equation}
\vec{w}_j(t+1)=\vec{w}_a(t)-\alpha \nabla L^{\mu_j(t)}(\vec{w}_a(t)),
\end{equation}
where $\alpha$ is the learning rate. 

{\bf (2) Decentralized Parallel SGD (DPSGD):} In the DPSGD algorithm \citep{dpsgd}, each learner $j$ computes the gradient at its own local weight $\vec{w}_j(t)$. The learning dynamics follows:
\begin{equation}
\vec{w}_j(t+1)=\vec{w}_{s,j}(t)-\alpha \nabla L^{\mu_j(t)}(\vec{w}_j(t)).
\end{equation}
where $\vec{w}_{s,j}(t)$ is the starting weight set to be the average weight of a subset of ``neighboring" learners of learner-$j$, which corresponds to the non-zero entries in the mixing matrix~\footnote{This is also called "gossip matrix" in the literature, e.g., \cite{koloskova2019decentralized}} defined in \citep{dpsgd} (note that $\vec{w}_{s,j}=\vec{w}_a$ if all learners are included as neighbors).

By averaging over all learners, the learning dynamics for the average weight $\vec{w}_a$ for both SSGD and DPSGD can be written formally the same way as:
\begin{equation}
\vec{w}_a(t+1)= \vec{w}_a(t)-\alpha \vec{g}_a,
\label{wa}
\end{equation}
where $\vec{g}_a =n^{-1}\sum_{j=1}^n \vec{g}_j$ is the average gradient and $\vec{g}_j$ is the gradient from learner-$j$. The difference between SSGD and DPSGD is the weight at which $\vec{g}_j$ is computed:  $\vec{g}_j\equiv \nabla L^{\mu_j(t)}(\vec{w}_a(t))$ is computed at $\vec{w}_a$ for SSGD; $\vec{g}_j\equiv \nabla L^{\mu_j(t)}(\vec{w}_j(t))$ is computed at $\vec{w}_j$ for DPSGD. The deviation of the weight for learner-$j$ from the average weight is defined as $\delta \vec{w}_j \equiv \vec{w}_j-\vec{w}_a$. It is easy to see that $\delta\vec{w}_j(t+1) = \vec{w}_{s,j}(t)-\vec{w}_a(t) -\alpha [\vec{g}_j(t)-\vec{g}_a(t)]$, which depends on gradients of the loss landscape. 


\subsection{Analysis from the Optimization Perspective}
The main difference between DPSGD and SSGD is that the stochastic gradients are calculated at different weights in DPSGD, while SSGD's stochastic gradient is calculated at the same weight.  Intuitively, DPSGD explores more space than SSGD, which may help explain the empirical success of DPSGD. We formalize this intuition into the following theorem, which shows that DPSGD is optimizing a smoother landscape than SSGD.

\begin{thm}
\label{thm:1}
Denote $\mathcal{F}_{t}$ by the filtration generated by all the random variables until the $t$-th iteration. Suppose $n$ is large enough that \small{$\left\|\frac{1}{n}\sum_{i=1}^{n}\nabla L^{\mu_i(t)}(\vec{w}_i(t))-\frac{1}{n-1}\sum_{i=1}^{n-1}\nabla L^{\mu_i(t)}(\vec{w}_i(t))\right\|\leq\epsilon$} 

\normalsize{almost surely, and assume $\delta \vec{w}_i(t)|\mathcal{F}_{t-1}\stackrel{i.i.d.}{\sim}\mathcal{N}(0,\sigma_w^2I)$ with $i=1,\ldots,n-1$. Then from the $(t-1)$-th iteration to $t$-th iteration, SSGD and DPSGD are doing one step of stochastic gradient descent on two different functions $L(\vec{w})$ and $\tilde{L}(\vec{w})\equiv \mathbb{E}_{\delta \vec{w}_i(t)}\left[L(\vec{w}+\delta\vec{w}_i(t))\,|\,\mathcal{F}_{t-1}\right]$, respectively. The DPSGD loss $\tilde{L}(\vec{w})$ is smoother than the SSGD loss $L(\vec{w})$ if $L(\vec{w})$ is Lipschitz continuous.}
\end{thm}
\textbf{Remark}: The proof of Theorem~\ref{thm:1} can be found in Appendix~\ref{appendix:theory1}. Here, we briefly mention its implications. A function $f$ is defined as $l_s$-smooth if $\|\nabla f(x)-\nabla f(y)\|\leq l_s\|x-y\|$ for any $x,y$, where $l_s$ is the smoothness parameter of $f$. The landscape of the function $f$ is smoother when $l_s$ is smaller. Assume $L(\vec{w})$ is $G$-Lipschitz continuous, by using Lemma 2 of~\cite{nesterov2017random}, we know that the DPSGD landscape 
$\tilde{L}(\vec{w})$ is $\frac{2G}{\sigma_w}$-smooth. According to the convergence theory of SGD and DPSGD for nonconvex functions~\cite{ghadimi2013stochastic,lian2017can,facebook-1hr}, the largest learning rate one can choose to guarantee convergence is $\frac{1}{l_s}$. For SSGD with the original loss landscape $L$, $l_s$ can be very large (even close to $+\infty$ due to the nonsmooth nature of the ReLU activation) while $l_s$ of the smoothed loss function 
$\tilde{L}$ for DPSGD is much smaller. This explains why we can use a larger learning rate in DPSGD as the landscape DPSGD sees has a smaller gradient-Lipschitz constant $l_s$ than that in SSGD. 

It is important to note that $l_s$ of the smoothed loss function $\tilde{L}$ in DPSGD depends on the standard deviation $\sigma_w$ of weights from different learners. Since $\sigma_w$ depends on the loss landscape and changes with time (see Fig.~2(b)), the smoothing effect in DPSGD is self-adjusting -- it is strong in the initial stage of training when the loss landscape is rough and becomes weaker as training progresses when the loss landscape becomes smoother. 
Our theoretical result suggests that this self-adjusting smoothing effect is responsible for DPSGD's convergence with a large learning rate in the large batch size setting. 
Next, we elaborate on this insight and verify it in a simple network for classification using MNIST dataset.

\mingrui{Note that the Theorem~\ref{thm:1} is only a one-step analysis. However it is not difficult to extend this insight to trajectory-based analysis. If we consider the perturbed objective $\tilde{L}(w)=\mathbb{E}_{\delta}\left[L(w+\delta)\right]$, where $\delta$ comes from the intrinsic noise by DPSGD, then we can utilize the descent lemma as shown in~\cite{ghadimi2013stochastic} to prove that DPSGD can converge to stationary point of $\tilde{L}(w)$ in polynomial time. However, without the inherent noise by DPSGD, the landscape is rough and that is the reason why SSGD diverges. SSGD may not be able to converge to the stationary point of $L(w)$ (since the large learning rate in large batch setting makes the descent lemma not applicable in this case) or $\tilde{L}(w)$ (since there is no noise and landscape-smoothing effect in SSGD, so SSGD does not optimize the smoothed landscape). This is also consistent with our empirical evidence.}
\subsection{The Landscape-dependent Self-Adjusting Learning Rate in DPSGD Help Convergence}

To understand the implication of the smoothing effect in DPSGD (Theorem 1) for learning dynamics, we define an effective learning rate $\alpha_e \equiv \alpha \vec{g}_a\cdot\vec{g}/||\vec{g}||^2$ by projecting the weight displacement vector $\Delta \vec{w}_a\equiv \alpha \vec{g}_a$ onto the direction of the gradient $\vec{g}\equiv \nabla L(\vec{w}_a)$ of the original loss function $L$ at $\vec{w}_a$. The learning dynamics, Eq.~\ref{wa}, can be rewritten as: 
\begin{equation}
\vec{w}_a(t+1)=\vec{w}_a(t)-\alpha_e \vec{g} +\vec{\eta}_{\bot},
\label{w_a}
\end{equation}
where the ``noise" term $\vec{\eta}_\bot \equiv -\alpha \vec{g}_a+\alpha_e \vec{g}$ describes the random weight dynamics in directions orthogonal to $\vec{g}$. The noise term has zero mean $\langle \vec{\eta}_\bot \rangle _\mu=0$ and the noise strength is characterized by its variance $\Delta(t)\equiv ||\vec{\eta}_\bot||^2$. 

\begin{figure}[htbp]
\centering
\includegraphics[width=1.0\linewidth]{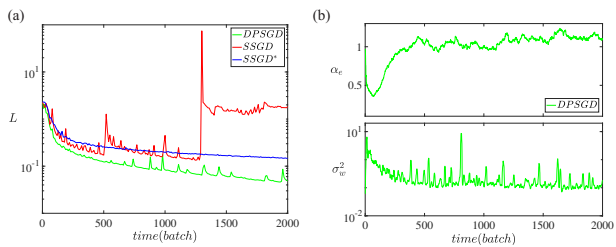}
\caption{ (a) Comparison of different multi-learner algorithms, DPSGD (green), SSGD (red), and SSGD$^*$ (blue) for a large learning rate $\alpha=1$. The adaptive learning rate allows DPSGD to converge while SSGD fails to converge. A fine-tuned SSGD$^*$ also converges but to an inferior solution. (b)  
The effective learning rate for DPSGD $\alpha_e(DPSGD)$ is self-adaptive to the landscape -- it is reduced in the beginning of training when gradients are large and recovers to $\sim \alpha$ when the gradients are small. The weight variance $\sigma_w^2 (t)$ has the opposite landscape-dependence as $\alpha_e$ and decreases with training time.} 
\label{conv} 
\vspace*{-0.03in}
\end{figure}

The effective learning rate $\alpha_e$ is related to the noise strength: $\alpha_e^2 =(\alpha^2 ||\vec{g}_a||^2-\Delta)/||\vec{g}||^2 $, which indicates that a higher noise strength $\Delta$ leads to a lower effective learning rate $\alpha_e$. 
The DPSGD noise $\Delta_{DP}$ is larger than the SSGD noise $\Delta_S$ by an additional noise term $\Delta^{(2)}(>0)$ that originates from the difference of local weights ($\vec{w}_j$) from their mean ($\vec{w}_a$): $\Delta_{DP}=\Delta_{S}+\Delta^{(2)}$, see \Cref{appendix:theory} for details.  By expanding $\Delta^{(2)}$ w.r.t. $\delta \vec{w}_j$, we obtain the average $\Delta^{(2)}$ over minibatch ensemble $\{\mu\}$:
\begin{small}
\begin{equation}
\begin{aligned}
  \langle \Delta^{(2)}\rangle_\mu &\equiv \alpha^2 \langle ||n^{-1}\sum_{j=1}^n[\nabla L^{\mu_j}(\vec{w}_j)- \nabla L^{\mu_j}(\vec{w}_a)]||^2\rangle_\mu\\
   & \approx \alpha^2 \sum_{k,l,l'} H _{kl}H_{kl'} C_{ll'},
    \label{D2}
\end{aligned}
\end{equation}
\end{small}
where $H_{kl}=\nabla^2 _{kl}L$ is the Hessian matrix of the loss function and $C_{ll'}=n^{-2}\sum_{j=1}^n \delta w_{j,l} \delta w_{j,l'}$ is the weight covariance matrix. From Eq.~\ref{D2} and the dependence of $\alpha_e$ on $\Delta$, it is clear that the effective learning rate in DPSGD depends directly on the loss landscape ($H$) and indirectly via the weight variance, $\sigma_w^2 =Tr(C)$, which decreases as the loss landscape becomes smooth (see Fig.~\ref{conv}(b)). 

It is important to stress that the noise $\vec{\eta}_\bot$ in Eq.\ref{w_a} is not an artificially added noise. It is intrinsic to the use of minibatches (random subsampling) in all SGD-based algorithms (including SSGD and DPSGD). The noise is increased in DPSGD due to the weight difference among different learners ($\delta \vec{w}_j$). The noise strength $\Delta$ varies in weight space via its dependence on the loss landscape, as explicitly shown in Eq.~\ref{D2}. However, besides its landscape dependence, SGD noise scales inversely with the minibatch size $B$~\citep{Chaudhari_2018}. With $n$ synchronized learners, the noise in SSGD scales as $1/(nB)$, which is too small to be effective for a large batch size $nB$. A main finding of our paper is that the additional landscape-dependent noise $\Delta^{(2)}$ in DPSGD can make up for the small SSGD noise when $nB$ is large and help enhance convergence in the large batch setting.

The landscape dependent smoothing effect in DPSGD (shown in Sec.~2.1) indicates that $\alpha_e$ in DPSGD is reduced at the beginning of training when the landscape is rough. To demonstrate effects of the landscape-dependent self-adjusting learning rates, we did detailed analysis in numerical experiments using the MNIST dataset.  In this experiment, we used $n=5$ learners with each learner a fully connected network with two hidden layers (50 units per layer) and we used $\vec{w}_{s,j}=\vec{w}_a$ for DPSGD. We focused on the large batch setting using $nB=2000$ and a large learning rate $\alpha = 1$. As shown in Fig.~\ref{conv}(a), DPSGD converges to a solution with a low loss ($2.1\%$ test error), but SSGD fails to converge. 

To understand the convergence in DPSGD, we computed the effective learning rate ($\alpha_e$) and the weight variance ($\sigma_w^2$) during training. As shown in Fig.~\ref{conv}(b) (upper panel), the effective learning rate $\alpha_e$ is reduced in DPSGD during early training ($0\le t\le 700$). This reduction of $\alpha_e$ is caused by the stronger noise $\Delta^{(2)}$ in DPSGD (see Fig.~\ref{Delta_2} in \Cref{appendix:theory}), which is essential for convergence when gradients are large in the beginning of the training process. In the later stage of the training process when gradients are smaller, the landscape-dependent DPSGD noise decreases and $\alpha_e$ {\em automatically}\/ increases back to be $\approx \alpha$ to allow fast convergence. From Eq.~\ref{D2}, the landscape-dependent noise in DPSGD depends on the weight variance. As shown in Fig.~\ref{conv}(b) (lower panel), the weight variance $\sigma^2_w $ has a time-dependent trend that is opposite to $\alpha_e$: $\sigma_w^2$ is large in the beginning of training when the landscape is rough and decreases as training progresses and the landscape becomes smoother. 

To show the importance of the landscape-dependent weight variance, we used SSGD$^*$, which injects a Gaussian noise with a constant variance to weights in SSGD, i.e., by setting $\delta \vec{w}_j \stackrel{i.i.d.}{\sim} \mathcal{N}(0,\sigma_0^2 I)$ with a constant $\sigma_0^2$. We found that SSGD$^*$ fails to converge for most choices of noise strength $\sigma_0^2$. Only by fine tuning $\sigma_0^2$ can SSGD$^*$ converge, but to an inferior solution with much higher loss and test error ($5.7\%$) as shown in Fig.~\ref{conv}(a). 

Finally, in addition to help convergence, we found that the landscape-dependent noise in DPSGD can also help find flat minima with better generalization in the large batch setting (see \Cref{appendix:flat} for details).  


\section{Experimental Methodology}
\label{sec:meth}

We implemented \sync and \dpsgd using PyTorch, OpenMPI, and NVidia NCCL. We ran experiments on a cluster of 8-V100-GPU x86 servers.  For CV tasks, we evaluated on \cifar (50,000 training samples, 178MB) and \imagenet (1.2 million training samples, 140GB). For ASR tasks, we evaluated on \swba (300 hours training data, 4,000,000 samples, 30GB) and \swbb (2000 hours training data, 30,000,000 samples, 216GB)\footnote{SWB-2000 is a more challenging task than ImageNet. It takes over 200 hours on 1 V100 GPU to finish training SWB-2000. SWB-2000 has 32,000 highly unevenly distributed classes, while ImageNet has 1000 evenly distributed classes.}.  We evaluate on 18 state-of-the-art NN models: 16 CNN models and 2 6-layer bi-directional LSTM models. We summarize the model size and training time in \Cref{tbl:model_size_training_time} of \Cref{appendix:meth}.
Also refer to \Cref{appendix:meth} for hardware configuration, software implementation, dataset and Neural Network (NN) model details.

\vspace*{-0.15in}
\section{Experimental Results}
\label{sec:results}
All the large batch experiments are conducted on 16 GPUs (learners). Batches are evenly distributed among learners, e.g., with sixteen learners, each learner uses a local batch size that is one sixteenth the overall batch size. A learner randomly picks a neighbor with which to exchange weights in each DPSGD iteration \citep{icassp20,lu2021optimal}. 
\subsection{\sync and \dpsgd Comparison on \cifar}
\textit{Single learner baseline} For \cifar experiments, we use the hyper-parameter setup proposed in \citep{pytorch-cifar}: a baseline 128 sample batch size and learning rate 0.1 for the first 160 epochs, learning rate 0.01 for the next 80 epochs, and learning rate 0.001 for the remaining 80 epochs. Using the same learning rate schedule, we keep increasing the batch size up to 8192. \Cref{tbl:cifar_baseline} in \Cref{appendix-results} records test accuracy under different batch sizes. Model accuracy consistently deteriorates beyond batch size 1024 because the learning rate is too small for the decreased number of parameter updates. 

\begin{table*}[t]
\centering
\small
\begin{tabular}{llllllllllll}
                &          & \tiny{Eff-B0}   & \tiny{SE-18} & \tiny{VGG}   & \tiny{Res-18}  & \tiny{Dense-121} & \tiny{Mobile} & \tiny{MobileV2} & \tiny{Shuffle} & \tiny{Google} & \tiny{ResNext-29} \\
                \hline
bs=128   & Baseline & 87.51 & 95.18 & 93.51 & 95.44 & 95.06  & 89.53   & 90.52     & 90.40    & 94.99   & 95.35   \\
lr=0.1                &          &       &       &       &       &        &         &           &          &         &         \\
                \hline
bs=1024  & SSGD    &\bf{91.92} & 94.52 & 93.12 & 94.59 & 95.11 & 92.24 & \bf{94.99} & 93.15 & \bf{95.32} & 95.42 \\             
lr=0.1                 &  DPSGD  & 91.69  & \bf{94.55} & \bf{93.15} & \bf{94.98} & \bf{95.12} & \bf{92.52} & 94.36 & \bf{93.55} & 95.18 & \bf{95.72} \\
\hline
bs=2048  & SSGD     & \bf{91.69} & 94.36 & 92.64 & \bf{94.96} & 95.11  & 91.72   & 94.24     & \bf{92.91}    & 94.76   & 94.19   \\
lr=0.2                & DPSGD    & 91.06 & \bf{94.70} & \bf{93.05} & 94.86 & \bf{95.32}  & \bf{92.72}   & \bf{94.51}     & 92.89    & \bf{94.80}   & \bf{95.30}   \\
\hline
bs=4096& SSGD     & \bf{91.62} & 94.28 & 92.68 & 94.30 & 94.72  & 91.68   & \bf{94.25}     & \bf{92.67}    & 94.36   & 93.21   \\
lr=0.4                 & DPSGD    & 91.23 & \bf{94.58} & \bf{92.72} & \bf{94.78} & \bf{95.24}  & \bf{92.03}   & 94.12     & 92.20    & \bf{94.99}   &  \bf{94.32}   \\
\hline
bs=8192 & SSGD     & 10    & 10    & 87.11 & 92.70 & 92.79  & 91.10   & \bf{93.22}     & 92.09    & 93.72   & 92.38   \\
lr=0.8                & DPSGD    & \bf{91.13} & \bf{90.48} & \bf{90.52} & \bf{94.34} & \bf{94.79}  & \bf{91.80}   & 93.09     & \bf{92.36}    & \bf{93.84}   & \bf{92.55}
\end{tabular}
\caption{DPSGD and SSGD comparison for \cifar, batch size 2048, 4096 and 8192, with learning rate set as 0.2, 0.4 and 0.8 respectively. All experiments are conducted on 16 GPUs (learners), with batch size per GPU 128, 256 and 512 respectively. Bold texts represent the best model accuracy achieved given the specific batch size and learning rate. When batch size is 8192, DPSGD significantly outperforms SSGD.  The batch size 128 baseline is presented for reference. bs stands for batch-size, lr stands for learning rate.}
\label{tab:cifar_comparison}
\end{table*}
\textit{DPSGD and SSGD Comparison} To improve model accuracy beyond batch size 1024, we apply the linear scaling rule (i.e., linearly increase learning rate w.r.t batch size)~\citep{resnet, facebook-1hr, zhang-ijcai-2016}. We use learning rate 0.1 for batch size 1024, 0.2 for batch size 2048, 0.4 for batch size 4096, and 0.8 for batch size 8192. \Cref{tab:cifar_comparison} compares \sync and DPSGD performance running with 16 GPUs (learners). \sync and \dpsgd perform comparably up to batch size 4096. When the batch size increases to 8192, DPSGD outperforms \sync in all but one case. Most noticeably, \sync diverges in \eff and \senet when the batch-size is 8192. \Cref{fig:cifar_large_bs} in \Cref{appendix-cifar10-progression} details the model accuracy progression versus epochs in each setting.
To better understand the loss landscape in \sync and \dpsgd training, we visualize the landscape with 2D contour projections and 2D Hessian projections in \Cref{sec:cifar-visualization}, using the method from  \citep{li-loss-nips}. Results in \Cref{sec:cifar-visualization} demonstrate that DPSGD can often find flatter optima than SSGD for \cifar tasks, which is consistent with results for MNIST shown in \Cref{appendix:flat}. 

\textit{Summary}  DPSGD outperforms SSGD for 9 out of 10 \cifar tasks in the large batch setting. Moreover, SSGD diverges on the \eff and \senet tasks. \dpsgd is more effective at avoiding early traps 
and reaching better solutions than \sync in the large batch setting.

\subsection{\sync and \dpsgd Comparison on ImageNet-1K}
\begin{table}[t]
\centering
\small
\begin{tabular}{llllllll}
                &          & AlexNet   & VGG & VGG-BN     \\
                \hline
bs=256   & Baseline & 56.31/79.05 &  69.02/88.66 & 70.65/89.92      \\
lr=1x  &          &  lr=0.01      &        &lr=0.1        \\
                \hline

bs=2048  & SSGD     & \bf{54.29/77.43} & \bf{67.67/87.91} & \bf{70.36/89.58}  \\
lr=8x                & DPSGD    & 53.71/76.91 & 67.28/87.58 & 69.76/89.31 \\
\hline
bs=4096& SSGD     & 0.10/0.50 & 0.10/0.50 & 65.39/86.51 \\
lr=16x                 & DPSGD    & \bf{52.53/76.01} & \bf{66.44/87.20} & \bf{68.86/88.82}   \\
\hline
bs=8192 & SSGD     & 0.10/0.50    & 0.10/0.50    & 0.10/0.50  \\
lr=32x                & DPSGD    & \bf{49.01/73.00} & \bf{65.00/86.11} & \bf{63.55/85.43} 
\end{tabular}
\caption{ImageNet-1K Top-1/Top-5 model accuracy (\%) comparison for batch size 2048, 4096 and 8192. All experiments are conducted on 16 GPUs (learners), with batch size per GPU 128, 256 and 512 respectively. Bold texts represent the best model accuracy achieved given the specific batch size and learning rate. The batch size 256 baseline is presented for reference. bs stands for batch-size, lr stands for learning rate. Baseline lr is set to 0.01 for AlexNet and VGG11, 0.1 for the other models. In the large batch setting, we use learning rate warmup and linear scaling as prescribed in \citep{facebook-1hr}. For rough loss landscape like AlexNet and VGG, \ssgd diverges when batch size is large whereas \dpsgd converges.}
\label{tab:imagenet_comparison}
\end{table}
\weiz{We test 6 CNN models -- AlexNet, VGG11, VGG11-BN, ResNet-50, ResNext-50 and DenseNet-161. Among them, AlexNet and VGG have rougher loss landscapes and can only work with smaller learning rates, while VGG11-BN, ResNet-50, ResNext-50, and DenseNet-161 have smoother loss landscapes thanks to the use of BatchNorm or Residual Connections, and thus can work with larger learning rates. We use the same baseline training recipe prescribed in \citep{pytorch-imagenet}: batch size 256, initial learning rate 0.01 for AlexNet and VGG-11 and 0.1 for the other 4 models, learning rate anneals by 0.1 every 30 epochs, 100 epochs in total. To study the model performance in the large batch setting, 
we follow the large batch size learning rate schedule prescribed in \citep{facebook-1hr}: learning rate warmup for the first 5 epochs and then learning rate linear scaling w.r.t batch size. For example, in the AlexNet batch-size 8192 experiment, the learning rate is gradually warmed-up from 0.01 to 0.32 in the first 5 epochs, annealed to 0.032 from epoch 31 to epoch 60, annealed to 0.0032 from epoch 61 to epoch 90, and annealed to 0.00032 from epoch 91 to epoch 100.
 \sync and \dpsgd achieve comparable model accuracy in the large batch setting (see \Cref{tab:imagenet_comparison_full} in \Cref{appendix-imagenet-progression}). Most noticeably, when batch-size increases to 8192, \sync diverges with AlexNet, VGG11, and VGG11-BN whereas \dpsgd converges as shown in \Cref{tab:imagenet_comparison}. \Cref{fig:imagenet_large_bs} in \Cref{appendix-imagenet-progression} details the model accuracy progression versus epochs in each setting.}
 
 \textit{Summary} For rough loss landscapes like AlexNet and VGG, \dpsgd converges whereas \sync diverges in the large batch setting.

\subsection{\sync and \dpsgd Comparison on ASR tasks}

\begin{table}
\begin{minipage}{0.5\textwidth}
\centering
\small
\begin{tabular}{|l|c|l|l|}
\hline
      & \multicolumn{3}{c|}{SWB-300}  \\ \hline
      & bs2048    & bs4096  & bs8192  \\ \hline
SSGD  & 1.58      & 10.37   & 10.37   \\ \hline
DPSGD & 1.59      & 1.60    & 1.66    \\ \hline
      & \multicolumn{3}{c|}{SWB-2000} \\ \hline
      & bs2048    & bs4096  & bs8192  \\ \hline
SSGD  & 1.46      & 1.46    & 10.37   \\ \hline
DPSGD & 1.45      & 1.47    & 1.47    \\ \hline
\end{tabular}
\caption{Heldout loss comparison for \sync and \dpsgd, evaluated on \swba and \swbb. There are 32000 classes in this task, a held-out loss 10.37 (i.e. $ln^{32000}$) indicates a complete divergence. bs stands for batch size.} 
\label{tab:asr_loss}
\end{minipage}
\vspace*{-0.1in}
\begin{minipage}{0.5\textwidth}
\centering
\includegraphics[scale=0.3]{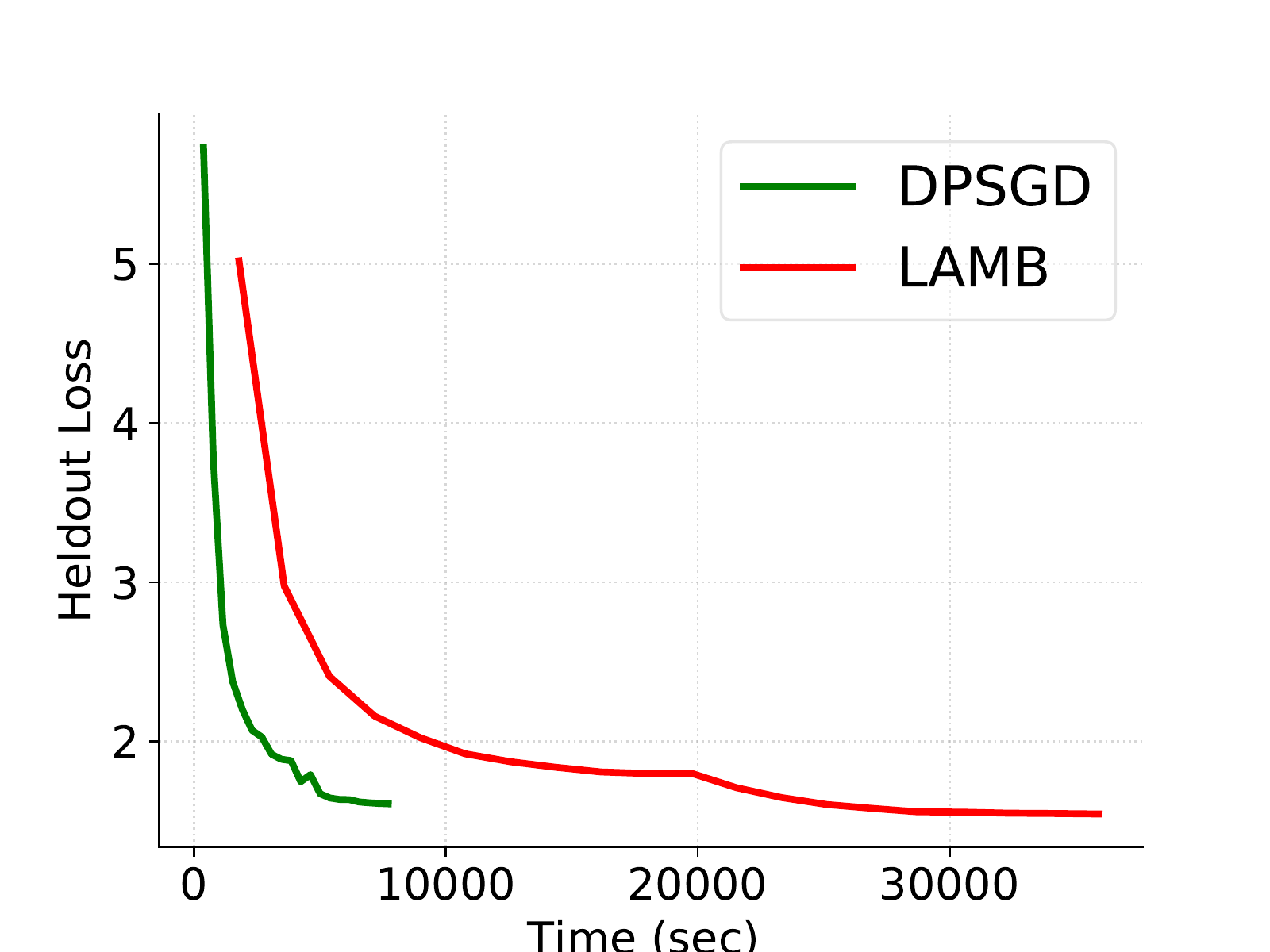}
\captionof{figure}{LAMB (a state-of-the-art SSGD based solution) and DPSGD comparison when there is a straggler that runs 5x slower than other learners in the system. SWB-300 task, batch size 4096, x-axis is running time and y-xais is the held-out loss.}
\label{fig:straggler}
\end{minipage}
\end{table}

Unlike CV tasks where CNNs and its residual connection variants are the dominant models, ASR tasks overwhelmingly adopt RNN/LSTM models that capture sequence features. Furthermore, Batch-Norm is known not to work well in RNN/LSTM tasks~\citep{rnn-bn}. Finally, there are over 32,000 different classes with wildy uneven distribution in our ASR tasks due to the Zipfian characteristics of natural language. All in all, ASR tasks present a much more challenging loss landscape than CV tasks to optimize over.

For the \swba and \swbb tasks, we follow the same learning rate schedule proposed in \citep{icassp19}: we use learning rate 0.1 for baseline batch size 256, and linearly warmup learning rate w.r.t the baseline batch size  for the first 10 epochs before annealing the learning rate by $\frac{1}{\sqrt{2}}$ for the remaining 10 epochs. For example, when using a batch size 2048, we linearly warmup the learning rate to 0.8 by the end of the 10th epoch before annealing. \Cref{tab:asr_loss} illustrates heldout loss for \swba and \swbb. In the \swba task, \sync diverges beyond batch size 2048 and \dpsgd converges well until batch size 8192. In the \swbb task, \sync diverges beyond batch size 4096 and \dpsgd converges well until batch size 8192.  \Cref{fig:asr_loss} in \Cref{appendix-swb-progression}  details the heldout loss progression versus epochs. 

\textit{Summary} For ASR tasks, SSGD diverges whereas DPSGD converges to baseline model accuracy in the large batch setting.

\subsection{Noise-injection and Learning Rate Tuning}
\label{sec:noise-lr-tuning}
In 7 out of 18 studied tasks, a large batch setting leads to a complete divergence in \sync: \eff, \senet, AlexNet, VGG11, VGG11-BN, \swba and \swbb. As discussed in \Cref{sec:theory}, the intrinsic landscape-dependent noise in \dpsgd effectively helps escape early traps (e.g., saddle points) and improves training by automatically adjusting learning rate.  In this section, we demonstrate these facts by systematically adding Gaussian noise (the same as the $SSGD^*$ algorithm in \Cref{sec:theory}) and decreasing the learning rate. We find that \sync might escape early traps but still results in a much inferior model compared to \dpsgd. 
\paragraph{Noise-injection}

In \Cref{fig:noise},
we systematically explore Gaussian noise injection with mean 0 and standard deviation (std) ranging from 10 to 0.00001 via binary search (i.e. roughly 20 configurations for each task). We found in the vast majority of the setups, noise-injection cannot escape early traps. In \eff, only when std is set to 0.04, does the model start to converge, but to a very low accuracy (test accuracy 22.15\% in \sync vs 91.13\% in \dpsgd). In \senet, when std is set to 0.01, the model converges to a reasonable accuracy (84.86\%) but still significantly lags behind its \dpsgd counterpart (90.48\%). In the \swba case, when std is 0.01, \sync shows an early sign of converging for the first 3 epochs before it starts to diverge. In the AlexNet, VGG11, VGG11-BN, and \swbb cases, we didn't find any configuration that can escape early traps. \Cref{fig:noise} characterizes our best-effort Gaussian noise tuning and its comparison against SSGD and DPSGD.
A plausible explanation is that Gaussian noise injection escapes saddle points very slowly, since Gaussian noise is 
isotropic and the complexity for finding local minima is dimension-dependent~\citep{ge2015escaping}. Deep Neural Networks are usually over-parameterized (i.e., high-dimensional), so it may take a long time to escape local traps. In contrast, the heightened landscape-dependent noise in DPSGD is anisotropic~\citep{Chaudhari_2018,YuTu} and can drive the system to escape in the right directions.
\paragraph{Learning Rate Tuning}

\begin{table}
  \small
  \centering
\begin{tabular}{lcccc}
                           &       & AlexNet  & VGG11  & VGG11-BN \\
\hline
  \multirow{2}{*}{lr$^*$=32x}    & \sync  & 0.10/0.50                       & 0.10/0.50                      & 0.10/0.50                        \\
  & DPSGD & 49.010/73.00                      & \bf{65.004/86.11}                      & 63.546/85.43                       \\
\hline
\multirow{2}{*}{lr=16x}   & \sync  & 0.10/0.50                      & 0.10/0.50                       & \bf{70.11/89.47}                      \\
& DPSGD &  \bf{49.26/73.14}                      & 62.046/83.98                     & 69.108/89.07                      \\
\hline
\multirow{2}{*}{lr=8x}  & \sync  & 46.40/70.25                      & 45.32/70.61                       & 69.54/89.22                       \\
& DPSGD & 47.78/71.89                      & 56.52/79.92                     &     68.98/88.78                   \\
\hline
\multirow{2}{*}{lr=4x} & \sync  & 41.77/66.44                     & 50.20/74.83                      & 68.61/88.57                       \\
                           & DPSGD & 42.18/66.96                      & 48.52/73.33                    & 67.98/88.22                      
\end{tabular}
\caption{ImageNet-1K learning rate tuning for AlexNet  VGG11, VGG11-BN with batch-size 8192. Bold text in each column indicates the best top-1/top-5 accuracy achieved across different learning rate and optimization method configurations for the corresponding batch size. \dpsgd consistently delivers the most accurate models.  
*The learning rate 1x used here corresponds to batch size 256 baseline learning rate, and we still adopt the same learning rate warmup, scaling and annealing schedule. Thus 32x refers to linear learning rate scaling when batch size is 8192. By reducing learning rate to 16x, 8x and 4x, \sync can escape early traps but still lags behind compared to \dpsgd in most cases.}
\vspace*{-0.15in}
\label{tab:imagenet_compare_lr}
\end{table}


\begin{table}
  \small
  \centering
\begin{tabular}{lcccc}
                           &       & {SWB-300} & SWB-300 & SWB-2000  \\
                           &       & (bs4096)  & (bs8192)&  (bs 8192) \\
\hline
  \multirow{2}{*}{lr$^*$=1.6/3.2}    & \sync  & 10.37                       & 10.37                      & 10.37                        \\
  & DPSGD & \bf{1.60}                      & \bf{1.66}                      & \bf{1.47}                       \\
\hline
\multirow{2}{*}{lr=0.8/1.6}   & \sync  & 10.37                      & 10.37                       & 10.37                       \\
& DPSGD & 1.65                      & 1.73                      & 1.48                       \\
\hline
\multirow{2}{*}{lr=0.4/0.8}  & \sync  & 1.76                      & 10.37                       & 1.51                       \\
& DPSGD & 1.77                      & 1.80                      & 1.52                       \\
\hline
\multirow{2}{*}{lr=0.2/0.4} & \sync  & 1.92                      & 2.05                      & 1.58                       \\
                           & DPSGD & 1.94                      & 2.00                      & 1.59                      
\end{tabular}
\caption{Decreasing learning rate for \swba and \swbb (bs stands for batch-size). Bold text in each column indicates the best held-out loss achieved across different learning rate and optimization method configurations for the corresponding batch size. \dpsgd consistently delivers the most accurate models. *learning rate 1.6 is used for bs4096 and learning rate 3.2 is used for bs8192.  We still adopt the same learning rate warmup, scaling and annealing schedule (baseline learning rate is 0.1 for batch size 256).}
\label{tab:swb_compare_lr}
\end{table}

\weiz{To make otherwise-divergent SSGD training converge in large batch setting, we systematically tune down the learning rates. \Cref{tab:imagenet_compare_lr} and \Cref{tab:swb_compare_lr} compare the model quality trained by SSGD and DPSGD of smaller learning rates in the large batch setting, for ImageNet and ASR tasks. \Cref{tab:cifar10_bs8192_compare_lr} in \Cref{appendix-cifar10-tuning} illustrates the similar learning rate tuning effort for \cifar tasks. As we can see, by using a smaller learning rate, \sync can escape early traps and converge, however it consistently lags behind \dpsgd in the large batch setting. Morever, DPSGD does not depend on such an exhaustive learning rate tuning to achieve convergence. DPSGD can simply follow the learning rate warm-up and linear scaling rules \cite{facebook-1hr} whereas SSGD is subject to much more stringent learning rate tuning. This implies DPSGD practitioners enjoy a much larger degree of freedom when it comes to hyper-parameter tuning in large batch setting than the SSGD practitioners.}

\textit{Summary} By systematically introducing landscape-independent noise and reducing the learning rate, \sync could escape early traps (e.g., saddle points), but results in much inferior models compared to \dpsgd in the large batch setting.
\subsection{End-to-End Run-time comparison}
We compare the end-to-end runtime between DPSGD and SSGD on both a low-latency network (e.g., HPC environment) and a high-latency network (e.g., Cloud environment). DPSGD consistently runs faster than SSGD. In the large batch setting where DPSGD converges better, DPSGD can leverage more computing hardware and parallelism to reach target accuracy, often several-fold faster than SSGD. Due to the page limit, we refer readers to Appendix~\ref{appendix:runtime} for the detailed experimental results and analysis.

In addition, DPSGD is immune to stragglers, while approaches that require global synchronization suffer slowdowns.  
\Cref{fig:straggler} demonstrates when there is a learner running 5x slower than other learners, DPSGD converges much faster than LAMB\cite{lamb}, a state-of-the art SSGD based large-batch training solution, on the SWB300 task. \weiz{This experiment demonstrates that even SSGD-variant algorithms (e.g., LAMB) can be designed to work for specific training tasks, DPSGD can simultaneously tackle the convergence problem and straggler-avoidance problem for the generic large batch training tasks.}
\section{Related Work}
\label{sec:related}
To increase parallelism in DDL, one must increase batch size, which often leads to a deteriorating model accuracy \citep{zhang2016icdm, tpu-mlperf}. Meticulous task-specific learning rate tuning for large batch training exists in CV training \citep{facebook-1hr, lars}, NLP training \citep{lamb} and ASR training \citep{icassp19}. Among them, layer-wise adaptive learning rate tuning schemes \citep{lars,lamb} rely on the Adam optimizer \citep{adam}, which may diverge on some simple convex functions \citep{adamsux}. 
In particular, \cite{lars,lamb} requires every learner to see other learner's gradients to calculate the large minibatch gradient, \cite{sam} optimizes both original loss function and the sharpness of the minimization, \cite{extragrad} calculates extra-gradient information and \cite{covnoise} leverages the covariance matrix of gradients noise. Furthermore, all above-mentioned approaches require global synchronization and suffer from the straggler problem: one slow learner can slow down the entire training process.


The noise in the stochastic gradient plays an important role in terms of generalization performance in deep learning. Keskar et al.~\citep{keskar2016large} show that large batch training procedures usually find sharp minima with poor generalization performance. This phenomenon is analyzed from different perspectives, including PAC-Bayesian learning theory~\citep{neyshabur2017exploring,neyshabur2017pac,dziugaite2017computing}, stochastic differential equation~\citep{jastrzkebski2017three}, Bayesian inference~\citep{smith2017bayesian} and optimization theory~\citep{kleinberg2018alternative}. There are several efforts trying to 
design algorithms to find flat minima that generalize better than SGD~\citep{chaudhari2016entropysgd,jastrzebski2018finding}. 


\section{Conclusion}
\label{sec:conclusion}

 \weiz{In practice, it is critical to turn around training and yield a reasonably accurate model in a short period of time. Thus, a DDL algorithm that can avoid divergence is highly desirable. SSGD, the de facto DDL algorithm, tends to diverge when learning rate must be large to compensate reduced number of parameter updates in the large batch setting. }
 
 In this paper, we find that in the large-batch and large-learning-rate setting, \dpsgd yields comparable model accuracy when \ssgd converges; moreover, \dpsgd converges when \ssgd diverges. 
We then investigate why \dpsgd outperforms SSGD for large batch training. Through detailed analysis on small-scale tasks and an extensive empirical study of a diverse set of modern DL tasks, we conclude that the landscape-dependent noise, which is strengthened in the \dpsgd system, self-adjusts the effective learning rate according to the loss landscape, helping convergence. \weiz{This self-adjusting learning rate effect is a mere by-product of the inherent loss-landscape-dependent-noise of the DPSGD training algorithm and requires no additional computation, no additional communication and no additional hyper-parameter tuning. } We provide both theoretical analysis and empirical evidence. Based on our findings, we recommend that DDL practitioners consider \dpsgd as an alternative when the batch size must be kept large, e.g., when a shorter run time to reach a reasonable solution is desired.


\newpage

\bibliography{refs}

\clearpage
\newpage
\appendix
\section{Proof of Theorem 1}
\label{appendix:theory1}

We first start to compare the learning dynamics of DPSGD and SSGD respectively.
For DPSGD, we have
\begin{equation}
	\label{eq:1}
	\vec{w}_a(t+1)=\vec{w}_a(t)-\alpha \cdot \frac{1}{n}\sum_{i=1}^{n}\nabla L^{\mu_i(t)}(\vec{w}_i(t)),
\end{equation}
where $n$ is the number of machines, $i=1,\ldots,n$ is the index of the machine, $\vec{w}_i(t)$ is the weight of the model at the $t$-th iteration on $i$-th machine, 
$\vec{w}_a(t)=\frac{1}{n}\sum_{i=1}^{n}\vec{w}_i(t)$, $L$ is the loss function, $\mu_i(t)$ denotes the minibatch sampled from the $i$-th machine at the $t$-th iteration, and $\alpha$ is the learning rate.  
In contrast, SSGD's update rule is
\begin{equation}
	\label{eq:2}
		\vec{w}_a(t+1)=\vec{w}_a(t)-\alpha \cdot \frac{1}{n}\sum_{i=1}^{n}\nabla L^{\mu_i(t)}(\vec{w}_a(t)).
\end{equation}

Define $\delta\vec{w}_i(t)=\vec{w}_a(t)-\vec{w}_i(t)$. 
Let us consider following fact: Given the realization of $\mu_i(t-1)$, $\vec{w}_i(t)$'s are mutually independent, and any $n-1$ random variables selected from $\{\delta\vec{w}_i(t)\}_{i=1}^{n}$ are mutually independent due to $\sum_{i=1}^{n}\delta \vec{w}_i(t)=0$.

When $n$ is sufficiently large, we have the surrogate minibatch gradient with batch size $n-1$ ($\frac{1}{n-1}\sum_{i=1}^{n-1}\nabla L^{\mu_i(t)}(\vec{w}_i(t))$) to be $\epsilon$-close to the minibatch gradient with size $n$ ($\frac{1}{n}\sum_{i=1}^{n}\nabla L^{\mu_i(t)}(\vec{w}_i(t))$), and hence can be regarded as approximate minibatch gradient with batch size $n-1$, which are sampled i.i.d. from $\{\delta\vec{w}_i(t)\}_{i=1}^{n-1}\,|\,\mathcal{F}_{t-1}$. Once we have the independence, we can find that both (\ref{eq:1}) and (\ref{eq:2}) are doing SGD update, with different objective functions. In addition, assuming $\{\delta\vec{w}_i(t)\}_{i=1}^{n-1}\,|\,\mathcal{F}_{t-1}
$ are i.i.d. Gaussian distribution is also reasonable due to the central limit theorem and the fact that $n$ is sufficiently large.


Then at the $t$-th iteration, (\ref{eq:1}) is using one step of SGD to optimize $L(\vec{w})$ directly, while (\ref{eq:2}) is using one step of SGD to optimize a smoothed version of $L$, which is $\mathbb{E}_{\delta \vec{w}_i(t)}\left[L(\vec{w}+\delta\vec{w}_i(t))\,|\,\mathcal{F}_{t-1}\right]$.

Suppose $L(\vec{w})$ is $G$-Lipschitz continuous, by using Lemma 2 of~\cite{nesterov2017random}, we know that the landscape DPSGD is trying to optimize over is 
$\tilde{L}(\vec{w})$ is $\frac{2G}{\sigma_w}$-smooth.


\section{Appendix for the Noise Analysis}
\label{appendix:theory}

To understand the origin of the noise term $\vec{\eta}$ in DPSGD, we decompose the gradient $\vec{g}_j$ for an individual learner-$j$:
\begin{eqnarray}
    \vec{g}_j&=&\vec{g}_0+\delta g_j^{(1)} + \delta g_j^{(2)} \nonumber \\&=&\nabla L^\mu (\vec{w}_a) +[\nabla L^{\mu_j} (\vec{w}_a)-\nabla L^\mu (\vec{w}_a)] \nonumber \\ &+& [\nabla L^{\mu_j} (\vec{w}_j)-\nabla L^{\mu_j} (\vec{w}_a)],
    \label{gj}
\end{eqnarray}
where the first term $\vec{g}_0\equiv \nabla L^\mu (\vec{w}_a)$ in the right hand side of Eq.~\ref{gj} is the gradient of the loss function over the ``superbatch" $\mu$ defined as the sum of all the minibatches for different learners at a given iteration: $\mu(t)=\sum_{j=1}^{n}\mu_j(t)$; the second term $\delta g_j^{(1)}\equiv \nabla L^{\mu_j} (\vec{w}_a)-\nabla L^\mu (\vec{w}_a)$ describes the gradient difference (fluctuation) between a minibatch $\mu_j$ and the superbatch $\mu$; the third term $\delta g_j^{(2)}\equiv \nabla L^{\mu_j} (\vec{w}_j)-\nabla L^{\mu_j} (\vec{w}_a)$ represents the difference (fluctuation) of the gradients at the individual weight $\vec{w}_j$ and at the average weight $\vec{w}_a$. Note that $\delta g_j^{(2)}=0$ in SSGD as the gradients are taken at the average weight $\vec{w}_a$ for all learners. By taking the average of Eq.~\ref{gj} over $j$, we have: $\vec{g}_a=\vec{g}_0+\delta g_a^{(1)}+\delta g_a^{(2)}$ with $\delta g_a^{(i)}=n^{-1}\sum_{j=1}^n \delta g_j^{(i)}$ ($i=1,2$). Here, $\delta g_a^{(1)}$ vanishes after averaging over all minibatch. $\delta g_a^{(0)}$ is due to superbatch-superbatch difference and $\delta g_a^{(2)}$ comes from weight-weight difference in DPSGD. The gradient fluctuation has zero mean and its variance given by: $\Delta ^{(2)}\equiv \alpha^2 ||\delta \vec{g}_a^{(2)}||^2 $. Finally, the noise strength in DPSGD $\Delta_{DP}$ can be expressed as:
\begin{equation}
    \Delta_{DP} \equiv ||\vec{\eta}||^2 =\Delta _ {S}+\Delta ^{(2)},  
\end{equation}
where $\Delta_{S}\equiv \alpha^2 (||\vec{g}_0||^2 -(\vec{g}_0\cdot \vec{g})^2/||\vec{g}||^2 )$ is the SSGD noise strength which is equivalent to the noise strength in a single-learner SGD algorithm with a superbatch (size $nB$).  
The $\Delta^{(2)}$ term only exists in DPSGD. In general, this additional contribution makes the learning noise larger in DPSGD than that in SSGD, although noise strength also depends on $\vec{g_a}$, $\vec{g}_0$, etc., which may be different for different algorithms.

\begin{figure}[htbp]
\centering
\includegraphics[width=0.5\linewidth]{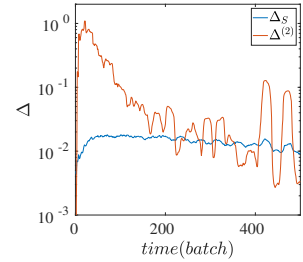}
\caption{ The noise in DPSGD can be decomposed into the SSGD noise $\Delta_S$ evaluated at the mean weight $\vec{w}_a$ plus an additional noise $\Delta^{(2)}(>0)$. The additional DPSGD noise $\Delta^{(2)}\gg \Delta_{S}$ in the beginning of the training before it decreases to become comparable to $\Delta_{S}$. }
\label{Delta_2}
\end{figure}

In Fig.~\ref{Delta_2}, we calculated these two noise components of DPSGD for the experiment shown in Fig.~\ref{conv}. Due to the large batch size we used in the experiment, $\Delta_{S}$ is very small during the training process. However, the additional landscape-dependent noise $\Delta^{(2)}$ in DPSGD can make up for the small SSGD noise when $nB$ is large and adaptively adjust the effectively learning rate $\alpha_e$ according to the loss landscape to help convergence. This additional landscape dependent noise in SGD is also responsible for finding flat minima with good generalization performance as shown in Fig.~\ref{conv_sup} in \Cref{appendix:flat}.

\section{Appendix for the effect of DPSGD noise in help finding flat minima with better generalization}  
\label{appendix:flat}
\begin{figure}[htbp]
\centering
\includegraphics[width=1.0\linewidth]{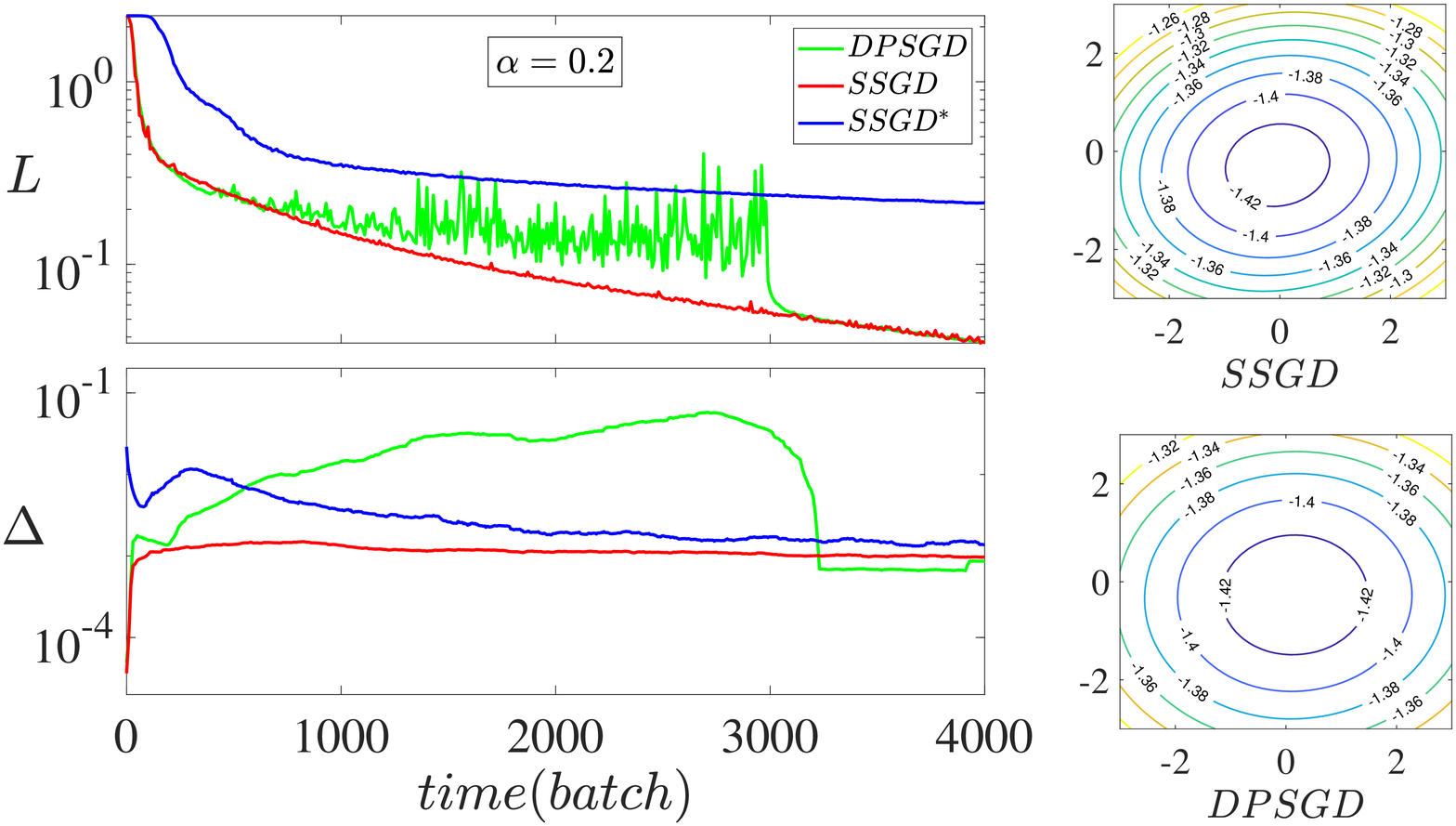}
\caption{ Comparison of different multi-learner algorithms, DPSGD (green), SSGD (red), and SSGD$^*$ (blue). For a smaller learning rate $\alpha=0.2$, both SSGD and DPSGD converge, however, DPSGD finds a flatter minimum with a lower test error than SSGD. The fixed noise SSGD$^*$ has the worst performance. See text for detailed description. }
\label{conv_sup} 
\end{figure}

To demonstrate the effect of the additional noise in DPSGD for finding flat minima, we consider a numerical experiment with a smaller learning rate $\alpha = 0.2$ for the MNIST dataset. We used $n=6$ and $\vec{w}_{s,j}(t)$ in DPSGD is the average weight of 2 neighbors on each side. In this case, both SSGD and DPSGD can converge to a solution, but their learning dynamics are different. As shown in Fig.~\ref{conv_sup} (upper panel), while the training loss $L$ of SSGD (red) decreases smoothly, the DPSGD training loss (green) fluctuates widely during the time window (1000-3000) when it stays significantly above the SSGD training loss. As shown in Fig.~\ref{conv_sup} (lower panel), these large fluctuations in $L$ are caused by the high and increasing noise level in DPSGD. This elevated noise level in DPSGD allows the algorithm to search in a wider region in weight space. At around time $3000$(batch), the DPSGD loss decreases suddenly and eventually converges to a solution with a similar training loss as SSGD. However, despite their similar final training loss, the DPSGD loss landscape is flatter (contour lines further apart) than SSGD landscape. Remarkably, the DPSGD solution has a lower test error (2.3$\%$) than the test error of the SSGD solution (2.6$\%$). We have also tried the SSGD$^*$ algorithm, but the performance ($3.9\%$ test error) is worse than both $SSGD$ and $DPSGD$. 


To understand their different generalization performance, we studied the loss function landscape around the SSGD and DPSGD solutions. The contour plots of the loss function $L$ around the two solutions are shown in the two right panels in Fig.~\ref{conv_sup}. We found that the loss landscape near the DPSGD solution is flatter than the landscape near the SSGD solution despite having the same minimum loss. Our observation is consistent with ~\citep{keskar2016large} where it was found that SSGD with a large batch size converges to a sharp minimum which does not generalize well. Our results are in general agreement with the current consensus that flatter minima have better generalization~\citep{Hinton1993Keeping,hochreiter1997flat,BaldassiE7655,chaudhari2016entropysgd,Zhang_2018}. 
It was recently suggested that the landscape-dependent noise in SGD-based algorithms can drive the system towards flat minima~\citep{YuTu}. However, in the large batch setting, the SSGD noise is too small to be effective. The additional landscape-dependent noise $\Delta^{(2)}$ in DPSGD, which 
also depends inversely on the flatness of the loss function (see Eq.~\ref{D2}), is thus critical for the system to find flatter minima in the large batch setting. 

\section{Appendix for Experimental Methodology}
\label{appendix:meth}
\begin{table*}
  \centering
  \small
  \begin{tabular}{|c|c|c|c|c|c|c|}
    \hline
         & \multicolumn{6}{|c|}{\normalsize{CIFAR10}}                                    \\
\hline
         & \eff & \senet    & \vgg     & \resnet & \densenet & \mobilenet \\
\hline
Size & 11.11 MB       & 42.95 MB    & 76.45 MB   & 42.63 MB   & 26.54 MB  & 12.27 MB   \\
\hline
Time & 2.92 Hr             & 1.58 Hr          & 1.08 Hr        & 1.37 Hr        & 5.48 Hr       & 1.02 Hr        \\
\hline
         & \multicolumn{4}{|c|}{\normalsize{CIFAR10}}                 & \normalsize{SWB300} & \normalsize{SWB2000}  \\
\hline
         & \mobilenetvtwo    & \shuf & \googlenet & \resnext & LSTM   & LSTM    \\
\hline
Size & 8.76 MB         & 4.82 MB     & 23.53 MB   & 34.82 MB   & 164.62 MB & 164.62 MB  \\
\hline
Time &   1.96 Hr           & 2.46 Hr         & 5.31 Hr        &   4.55 Hr      & 26.88 Hr  & 203.21 Hr \\
\hline
        & \multicolumn{6}{|c|}{\normalsize{ImageNet-1K}} \\
\hline
        & AlexNet   & VGG   & VGG-BN & ResNet-50  & ResNext-50 & DenseNet-161   \\
\hline
Size    & 233.08 MB  & 506.83 MB & 506.85 MB & 97.49 MB     & 95.48 MB     & 109.41 MB  \\
\hline
Time    & 190.67 Hr  & 168.67 Hr & 204.27 Hr  & 238.8 Hr    & 341.33 Hr    & 664.53 Hr \\
\hline
\end{tabular}
	\caption{Evaluated workload model size and training time. Training time is measured when running on 1 V100 GPU. \cifar is trained with batch size 128 for 320 epochs. \imagenet is trained with batch size 256 for 100 epochs. \swba and \swbb are trained with batch size 128 for 16 epochs.}
	\label{tbl:model_size_training_time}
\end{table*}
\subsection{Software and Hardware}
We use PyTorch 1.1.0 (Torchvision 0.2.0) as the single learner DL engine. Our communication library is built with CUDA 9.2 compiler, the CUDA-aware OpenMPI 3.1.1, and g++ 4.8.5 compiler. Concurrency control of computation threads and communication threads is implemented via Pthreads.
We run our experiments on a cluster of 8-V100 GPU servers. Each server has 2 sockets and 9 cores per socket. Each core is an Intel Xeon E5-2697 2.3GHz processor. Each server is equipped with 1TB main memory and 8 V100 GPUs. Between servers are 100Gbit/s Ethernet connections. GPUs and CPUs are connected via PCIe Gen3 bus, which has a 16GB/s peak bandwidth in each direction per socket.

\subsection{Dataset and Models}
	

We evaluate on two types of DL tasks: CV and ASR. For CV task, we evaluate on \cifar dataset~\citep{krizhevsky2009learning}, which comprises of a total of 60,000 RGB images of size 32 $\times$ 32  pixels partitioned into the training set (50,000 images) and the test set (10,000 images) and \imagenet dataset ~\citep{imagenet}, which comprises of 1.2 million training images (256x256 pixels) and 50,000 (256x256 pixels) testing images. We test \cifar with 10 representative CNN models ~\citep{pytorch-cifar}. The 10 CNN models are: (1) \eff , with a compound coefficient 0 in the basic EfficientNet architecture ~\citep{efficientnet}. (2) \senet, which stacks Squeeze-and-Excitation blocks ~\citep{senet} on top of a ResNet-18 model. (3) \vgg, a 19 layer instantiation of VGG architecture \citep{vgg}. (4) \resnet, a 18 layer instantiation of ResNet architecture \citep{resnet}. (5) \densenet, a 121 layer instantiation of DenseNet architecture \citep{densenet}. (6) \mobilenet, a 28 layer instantiation of MobileNet architecture \citep{mobilenet}. (7) \mobilenetvtwo, a 19 layer instantiation of \citep{mobilenetv2} architecture that improves over MobileNet by introducing linear bottlenecks and inverted residual block. (8) \shuf, a 50 layer instantiation of ShuffleNet architecture \citep{shufflenet}. (9) \googlenet, a 22 layer instantiation of Inception architecture \citep{googlenet}. (10) \resnext, a 29 layer instantiation of \citep{resnext} with bottlenecks width 64 and 2 sets of aggregated transformations. The detailed model implementation refers to \citep{pytorch-cifar}. Among these models, ShuffleNet, MobileNet, MobileNet-V2, EfficientNet represent the low memory footprint models that are widely used on mobile devices, where federated learnings is often used. The other models are standard CNN models that aim for high accuracy. We test 6 CNN models for \imagenet, AlexNet \citep{alexnet}, VGG11 \citep{vgg}, VGG11 with BatchNorm \citep{batchnorm} VGG11-BN, ResNet-50 \citep{resnet}, ResNext-50 \citep{resnext}, and DenseNet-161 \citep{densenet}.  

For ASR tasks, we evaluate on \swba and \swbb dataset. The input feature (i.e. training sample) is a fusion of FMLLR (40-dim), i-Vector (100-dim), and logmel with its delta and double delta (40-dim $\times$3). \swba, whose size is 30GB,  contains roughly 300 hour training data of over 4 million samples. \swbb, whose size is 216GB, contains roughly 2000 hour training data of over 30 million samples. The size of \swba held-out data is 0.6GB and the size of \swbb held-out data is 1.2GB. The acoustic model is a long short-term memory (LSTM) model with 6 bi-directional layers. Each layer contains 1,024 cells (512 cells in each direction). On top of the LSTM layers, there is a linear projection layer with 256 hidden units, followed by a softmax output layer with 32,000 (i.e. 32,000 classes) units corresponding to context-dependent HMM states. The LSTM is unrolled with 21 frames and trained with non-overlapping feature subsequences of that length.  This model contains over 43 million parameters and is about 165MB large.

\Cref{tbl:model_size_training_time} summarizes the model size and training time (on 1 V100 GPU) for evaluated tasks. \cifar tasks train 320 epochs, \imagenet tasks train 100 epochs, and all ASR tasks train 16 epochs. 


\section{Appendix for Results Section}
\label{appendix-results}
\subsection{CIFAR-10 Hyper-Parameter Tuning}
\label{appendix-cifar10-tuning}
\begin{table*}[t]
  \centering
  \small
\begin{tabular}{llllllll}
\hline
               & \multicolumn{7}{c}{Batch Size}                                                                                                                                                                \\ \hline
               & \multicolumn{1}{c}{128} & \multicolumn{1}{c}{256} & \multicolumn{1}{c}{512} & \multicolumn{1}{c}{1024} & \multicolumn{1}{c}{2048} & \multicolumn{1}{c}{4096} & \multicolumn{1}{c}{8192} \\ \hline
\eff           & 87.51                    & 89.32                    &  91.28                   & \bf{91.92}                     & 90.62                     & 88.00                     & 84.85                     \\ \hline
\senet         & \bf{95.18}                    & 94.84                    &  94.83                   & 94.52                     & 93.83                     & 92.94                     & 91.69                     \\ \hline
\vgg           & 93.51                    & \bf{93.78}                    &  93.35                   & 93.12                     & 92.64                     & 91.82                     & 87.76                     \\ \hline
\resnet        & \bf{95.44}                    & 95.26                    &  95.08                   & 94.59                     & 94.96                     & 92.98                     & 91.24                     \\ \hline
\densenet      & 95.06                    & 95.27                    &  \bf{95.42}                   & 95.11                     & 94.81                     & 93.09                     & 92.34                     \\ \hline
\mobilenet     & 89.53                    & 90.96                    &  \bf{92.39}                   & 92.24                     & 91.22                     & 89.54                     & 86.59                     \\ \hline
\mobilenetvtwo & 90.52                    & 92.93                    &  94.17                   & \bf{94.99}                     & 93.71                     & 91.97                     & 89.81                     \\ \hline
\shuf          & 90.4                     & 92.27                    &  92.82                   & \bf{93.15}                     & 91.94                     & 90.59                     & 87.81                     \\ \hline
\googlenet     & 94.99                    & 95.06                    &  94.97                   & \bf{95.32}                     & 94.05                     & 92.78                     & 91.09                     \\ \hline
\resnext       & 95.35                    & \bf{95.66}                    &  95.31                   & 95.42                     & 94.24                     & 93.00                     & 91.06                     \\ \hline
\end{tabular}
\caption{\cifar accuracy (\%) with different batch size. Across runs, learning rate is set as 0.1 for first 160 epochs, 0.01 for the next 80 epochs and 0.001 for the  last 80 epochs. Model accuracy consistently deteriorates when batch size is over 1024. Bold text in each row represents the highest accuracy achieved for the corresponding model, e.g., \eff achieves highest accuracy at 91.92\% with batch size 1024. }
\vspace*{-0.15in}
\label{tbl:cifar_baseline}
\end{table*}
For \cifar experiments, we use the hyper-parameter setup proposed in \citep{pytorch-cifar}: a baseline 128 sample batch size and learning rate 0.1 for the first 160 epochs, learning rate 0.01 for the next 80 epochs, and learning rate 0.001 for the remaining 80 epochs. Using the same learning rate schedule, we keep increasing the batch size up to 8192. \Cref{tbl:cifar_baseline} records test accuracy under different batch sizes. Model accuracy consistently deteriorates beyond batch size 1024 because the learning rate is too small for the decreased number of parameter updates.

\begin{table*}
  \small
  \centering
\begin{tabular}{llllllllllll}
  &       & Eff-B0              & SE-18                     & VGG                        & Res-18                    & Dense-121                    & Mobile                 & MobileV2                 & Shuffle                  & Google                   & ResNext-29                     \\
  \hline
\multirow{2}{*}{lr=0.8} & SSGD  & 10.00                       & 10.00                       & 87.11                       & 92.7                        & 92.79                       & 91.10                       & \bf{93.22} & 92.09                       & 93.72                       & 92.38                       \\
& DPSGD & \bf{91.13} & 90.48                       & 90.52                       & \bf{94.34} & \bf{94.79} & \bf{91.80} & 93.09                       & \bf{92.36} & \bf{93.84} & 92.55
\\
\hline
\multirow{2}{*}{lr=0.4} & SSGD  & 88.61                       & 92.84                       & 91.06                       & 91.98                       & 93.42                       & 91.13                       & 93.11                       & 91.54                       & 92.85                       & 89.70                       \\
                        & DPSGD & 89.80                       & \bf{94.00} & \bf{91.93} & 93.91                       & 94.32                       & 91.38                       & 93.14                       & 91.68                       & 93.49                       & \bf{92.79} \\
\hline
\multirow{2}{*}{lr=0.2} & SSGD  & 88.03                       & 92.41                       & 90.51                       & 92.13                       & 92.98                       & 88.38                       & 91.68                       & 90.14                       & 92.44                       & 91.31                       \\
                        & DPSGD & 87.69                       & 93.11                       & 91.59                       & 93.30                       & 94.28                       & 89.18                       & 92.52                       & 90.13                       & 93.41                       & 91.79                      

\end{tabular}
\caption{\cifar with batch size 8192. By reducing learning rate, \sync can escape early traps but still lags behind \dpsgd. Bold text in each column indicates the best accuracy achieved for that model across different learning rate and optimization method configurations. \dpsgd consistently delivers the most accurate models.}
\label{tab:cifar10_bs8192_compare_lr}
\end{table*}

\subsection{CIFAR-10 Training Progression}
\label{appendix-cifar10-progression}
\begin{figure*}
\centering
    \subfloat[\cifar convergence, bs=1024, lr=0.1]{
    {\includegraphics[width=2.0\columnwidth]{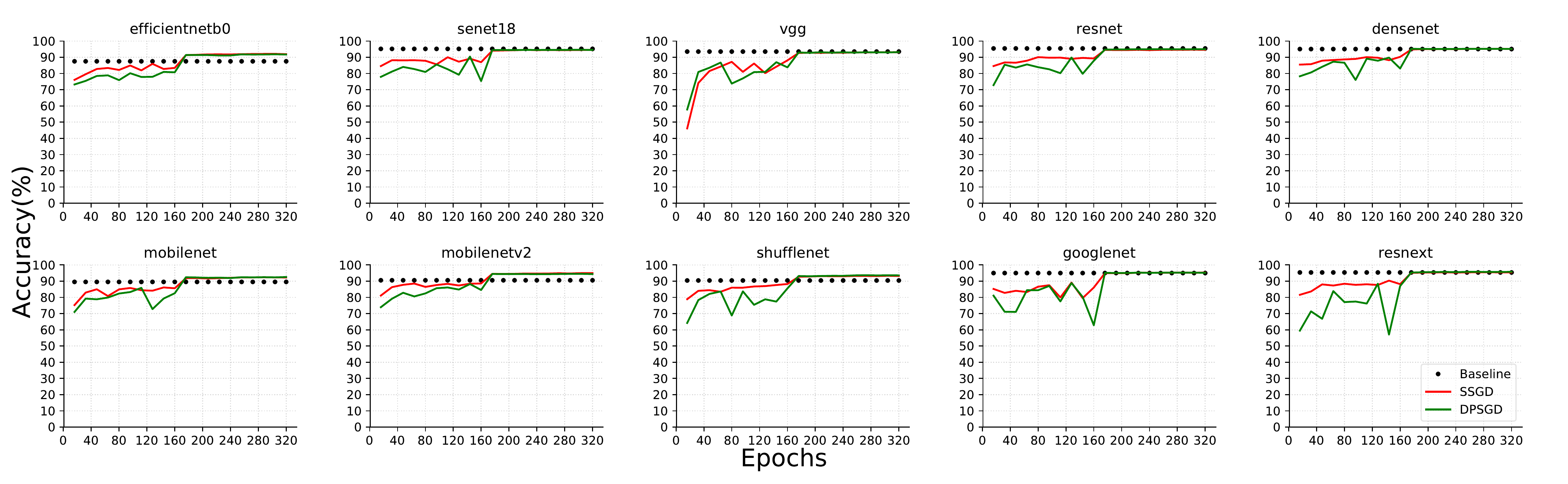}   
    }\label{fig:cifar-convergence-bs1024}
  }
  \\
  \subfloat[\cifar convergence, bs=2048, lr=0.2]{
    {\includegraphics[width=2.0\columnwidth]{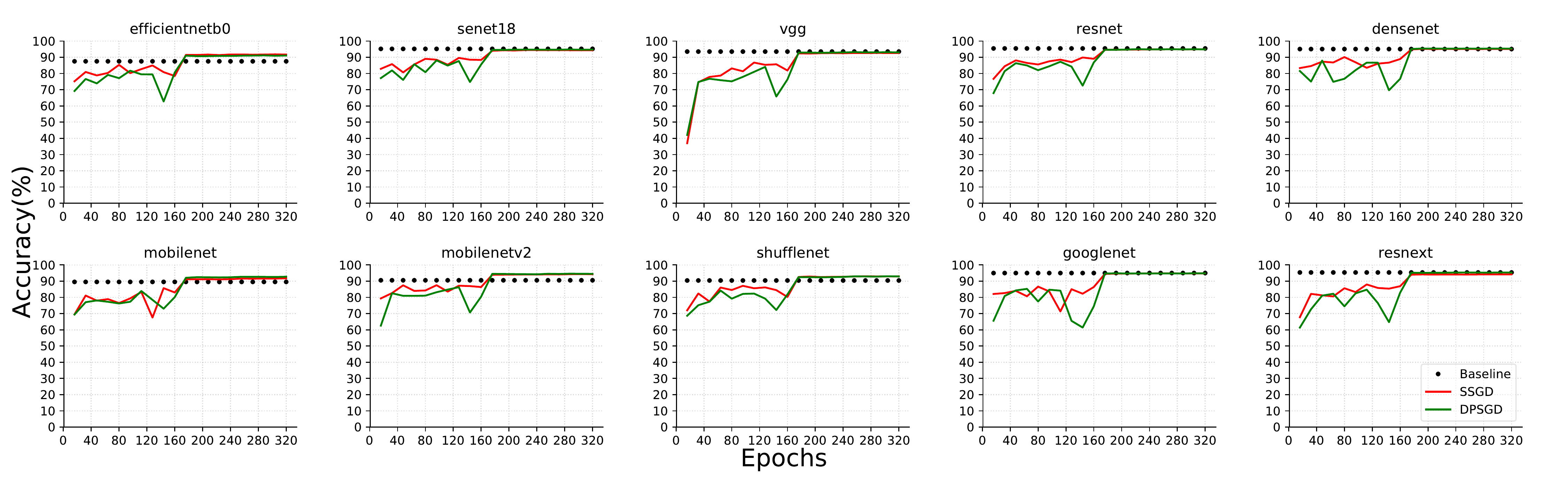}   
    }\label{fig:cifar-convergence-bs2048}
  }
  \\
  \subfloat[\cifar convergence, bs=4096, lr=0.4]{
    {\includegraphics[width=2.0\columnwidth]{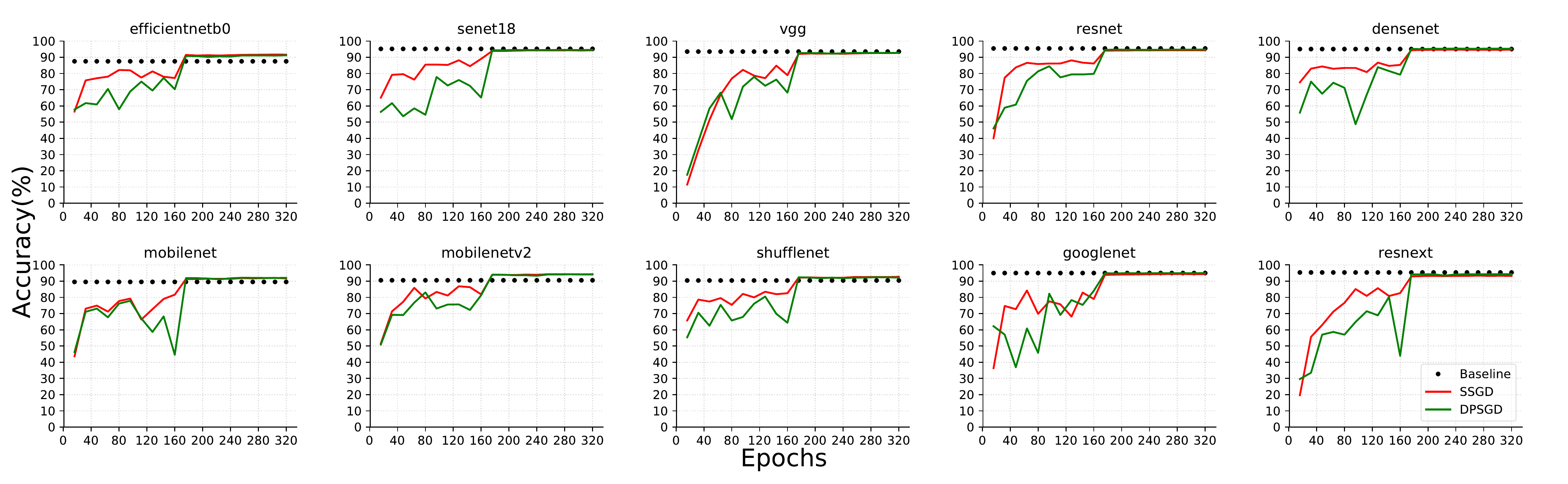}   
    }\label{fig:cifar-convergence-bs4096}
  }
  \\
  \subfloat[\cifar convergence, bs=8192, lr=0.8]{
    {\includegraphics[width=2.0\columnwidth]{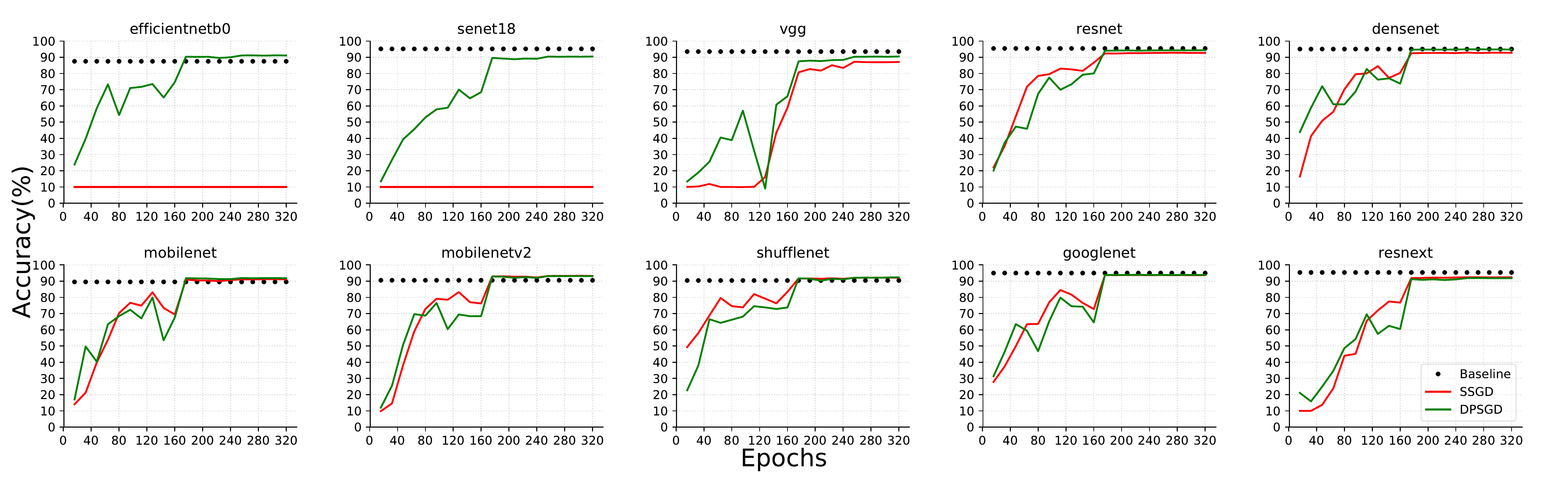}    
    }
    \label{fig:cifar-convergence-bs8192}
  }
  \caption{CIFAR-10 SSGD DPSGD comparison for batch size 2048, 4096 and 8192, with learning rate set as 0.2, 0.4 and 0.8 respectively. All experiments are conducted on 16 GPUs (learners), with batch size per GPU 128,256 and 512 respectively. When batch size is 8192, \dpsgd significantly outperforms \sync. bs stands for batch-size, lr stands for learning rate. The dotted black line represents the bs=128 baseline. }
  \label{fig:cifar_large_bs}
\end{figure*}

\Cref{fig:cifar_large_bs} illustrates \sync and \dpsgd comparison for \cifar. \sync and \dpsgd perform comparably up to batch size 4096. When batch size increases up to 8192,
\dpsgd outperforms \sync in all but one cases. Noticeably, \sync diverges in \eff and \senet when batch-size is 8192.

\subsection{\cifar Loss Landscape Visualization}
\label{sec:cifar-visualization}
\begin{figure*}[t]
\small
\centering

  \subfloat[VGG-S]{
      {\includegraphics[scale=0.12]{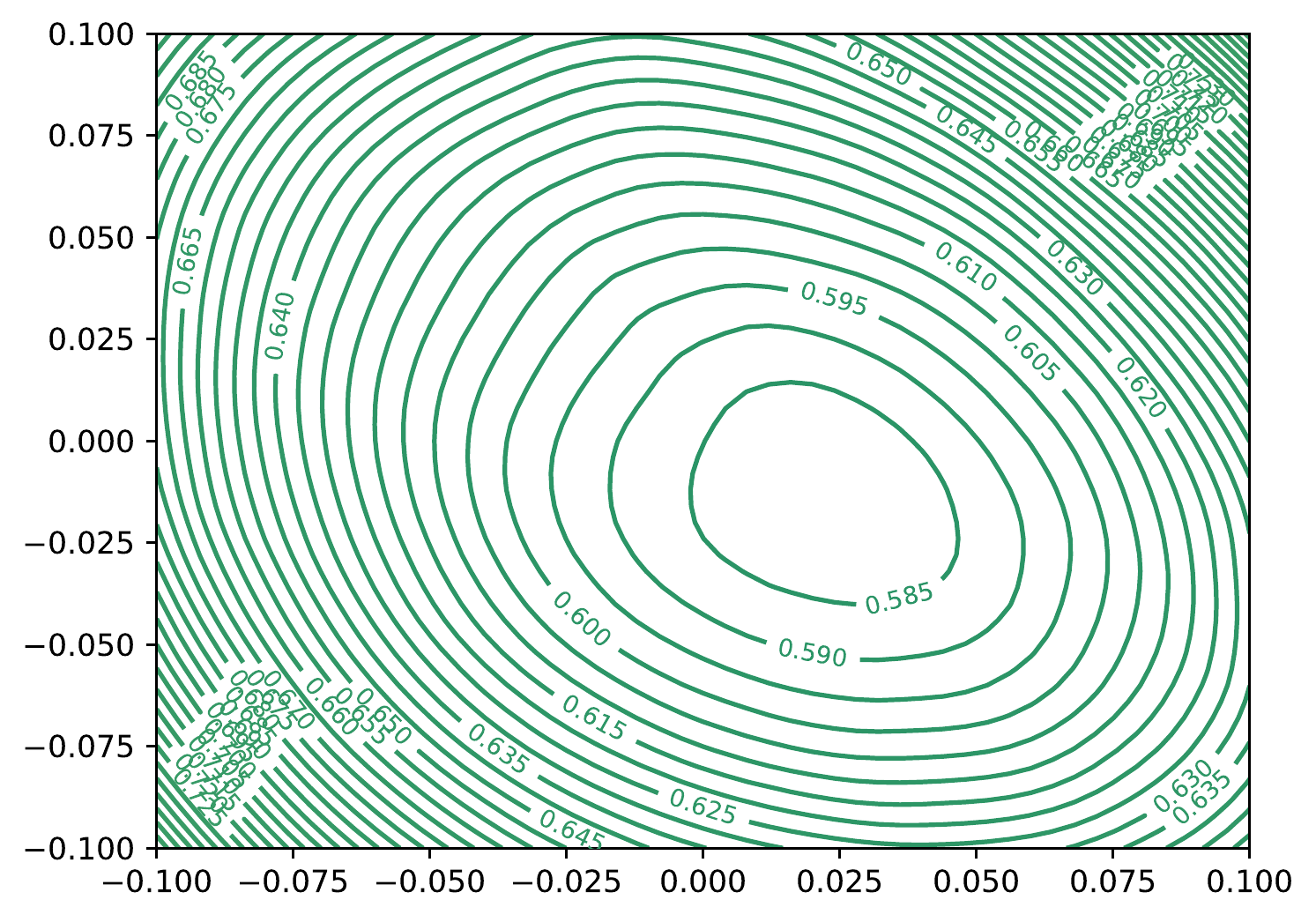}
      }\label{fig:vgg-sync-2d}
    }
    \subfloat[VGG-DP]{
      {\includegraphics[scale=0.12]{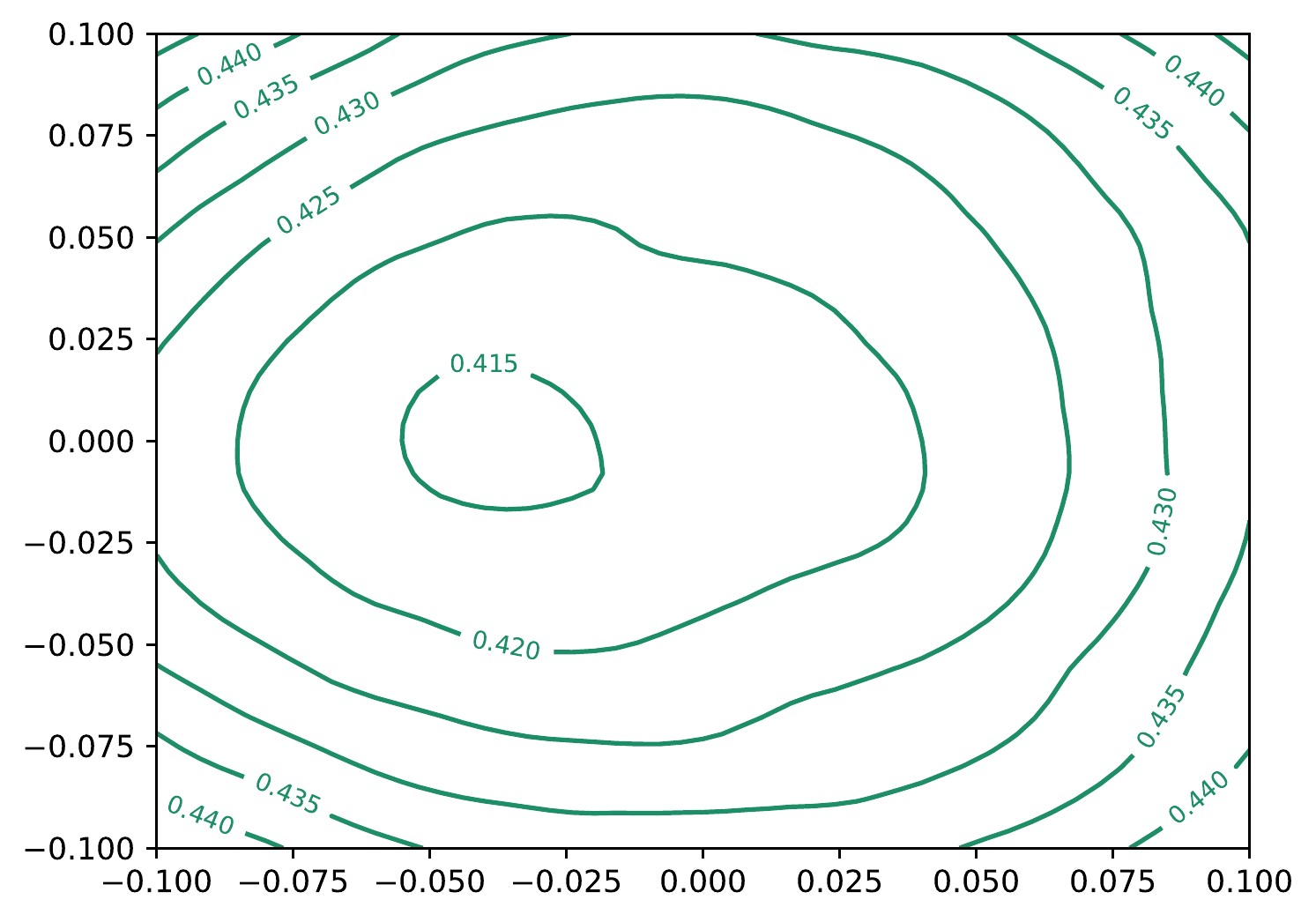}
      }\label{fig:vgg-dpsgd-2d}
    }
  \subfloat[ResN-S]{
      {\includegraphics[scale=0.12]{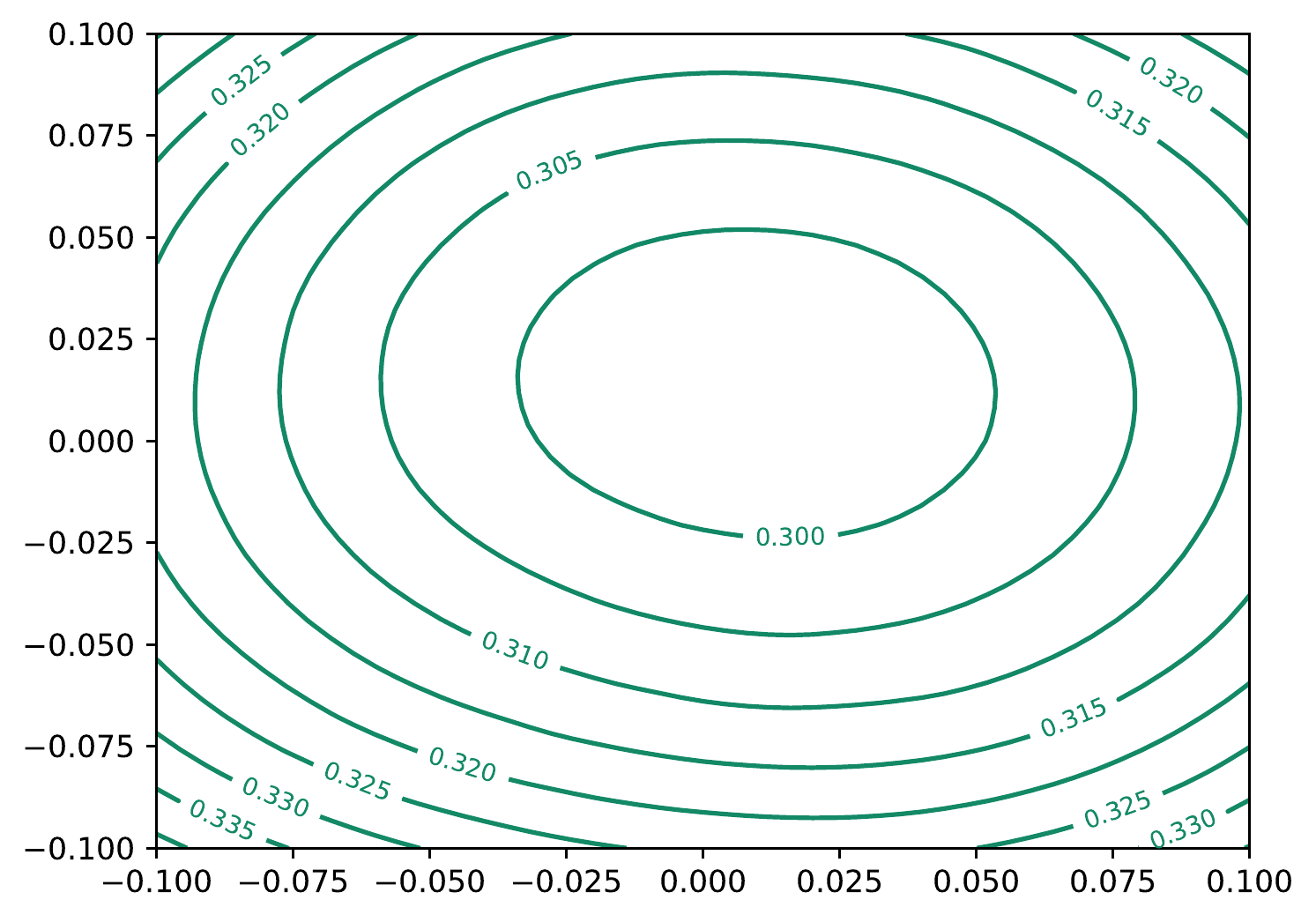}
      }\label{fig:resnet-sync-2d}
    }
    \subfloat[ResN-DP]{
      {\includegraphics[scale=0.12]{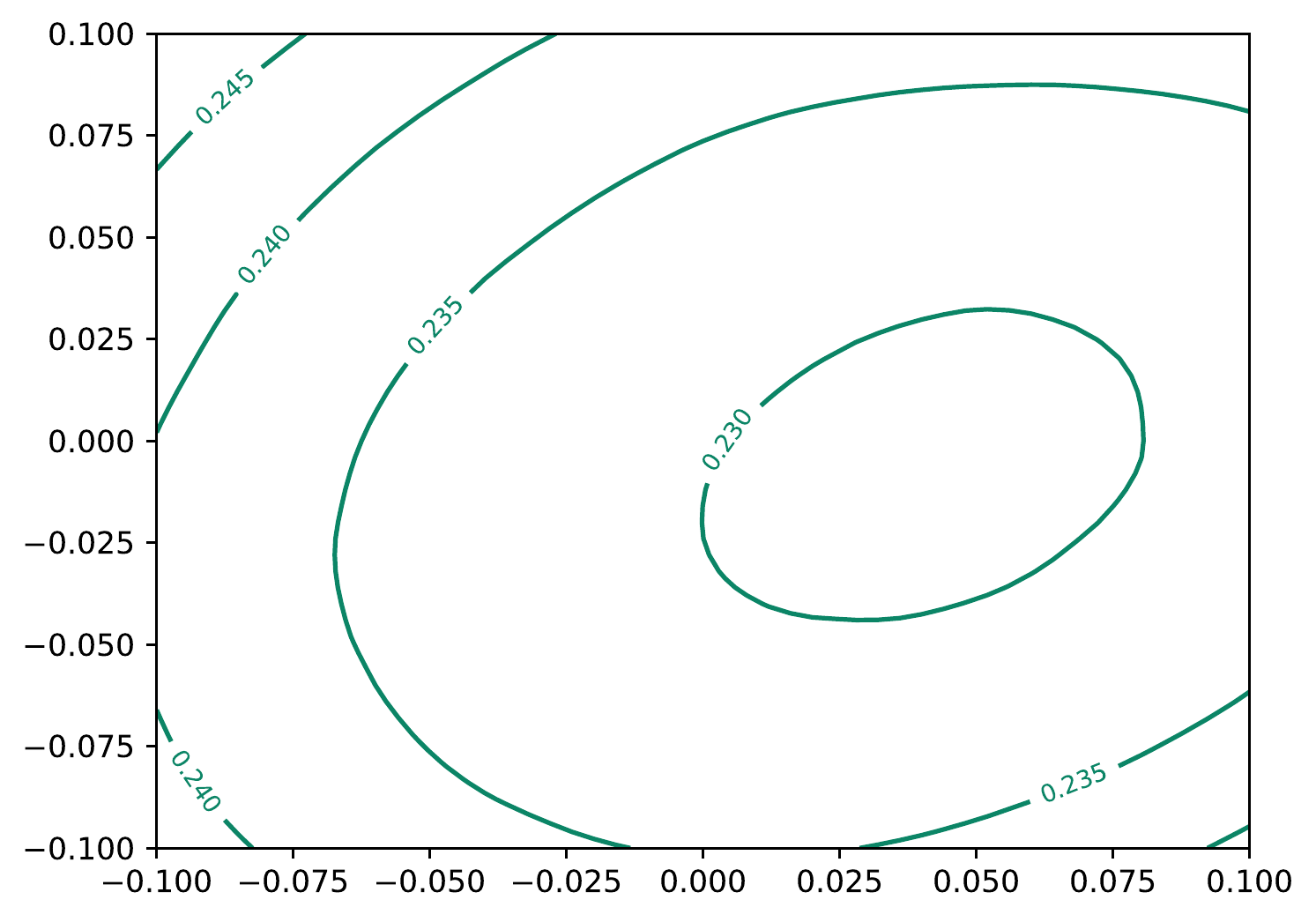}
      }\label{fig:resnet-dpsgd-2d}
    }
    \subfloat[DenseN-S]{
      {\includegraphics[scale=0.12]{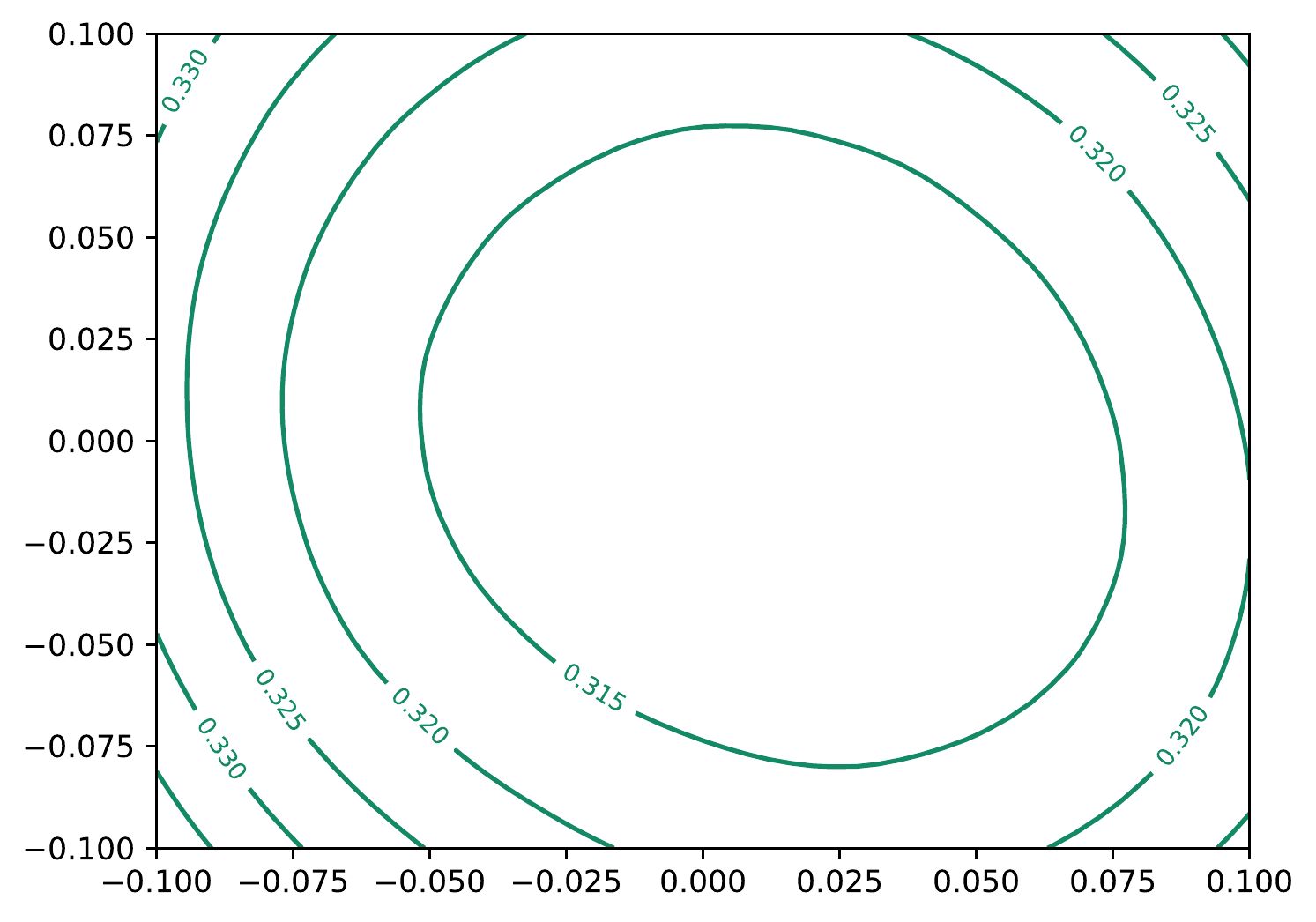}
      }\label{fig:densenet-sync-2d}
    }
    \subfloat[DenseN-DP]{
      {\includegraphics[scale=0.12]{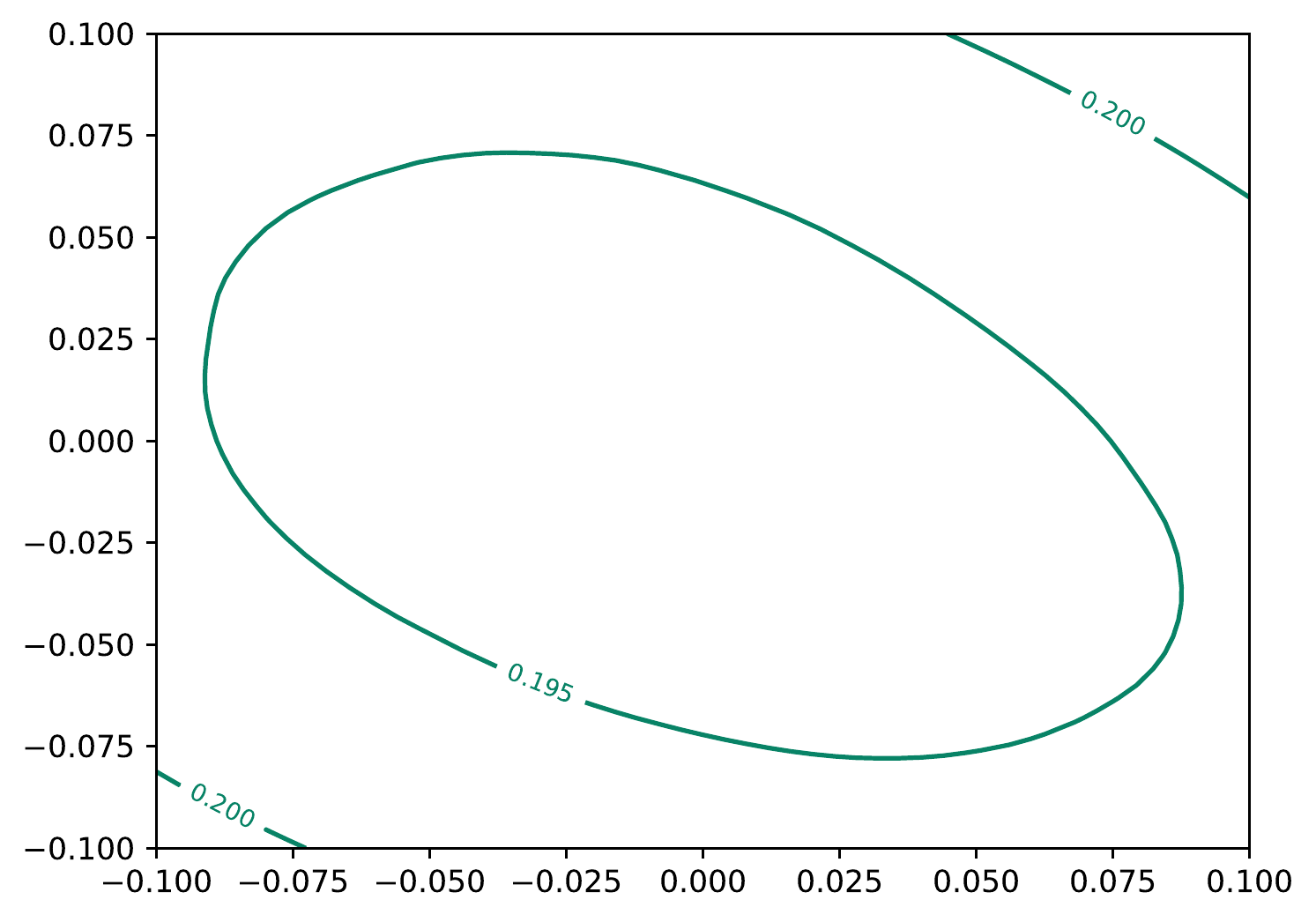}
      }\label{fig:densenet-dpsgd-2d}
    }
\vspace*{-0.1in}
\caption{\cifar 2D contour plot. The more widely spaced contours represent a flatter loss landscape and a more generalizable solution. The distance between each contour line is 0.005 across all the plots. We plot against the model trained at the end of 320th epoch. VGG: \vgg, ResN: \resnet, DenseN: \densenet, -S: -\sync, -DP: -\dpsgd}
\vspace*{-0.2in}
    \label{fig:cifar_2dcontour}
\end{figure*}

\begin{figure*}[t]
\small
    \centering
    \subfloat[{VGG-S}]{
      {\includegraphics[scale=0.16]{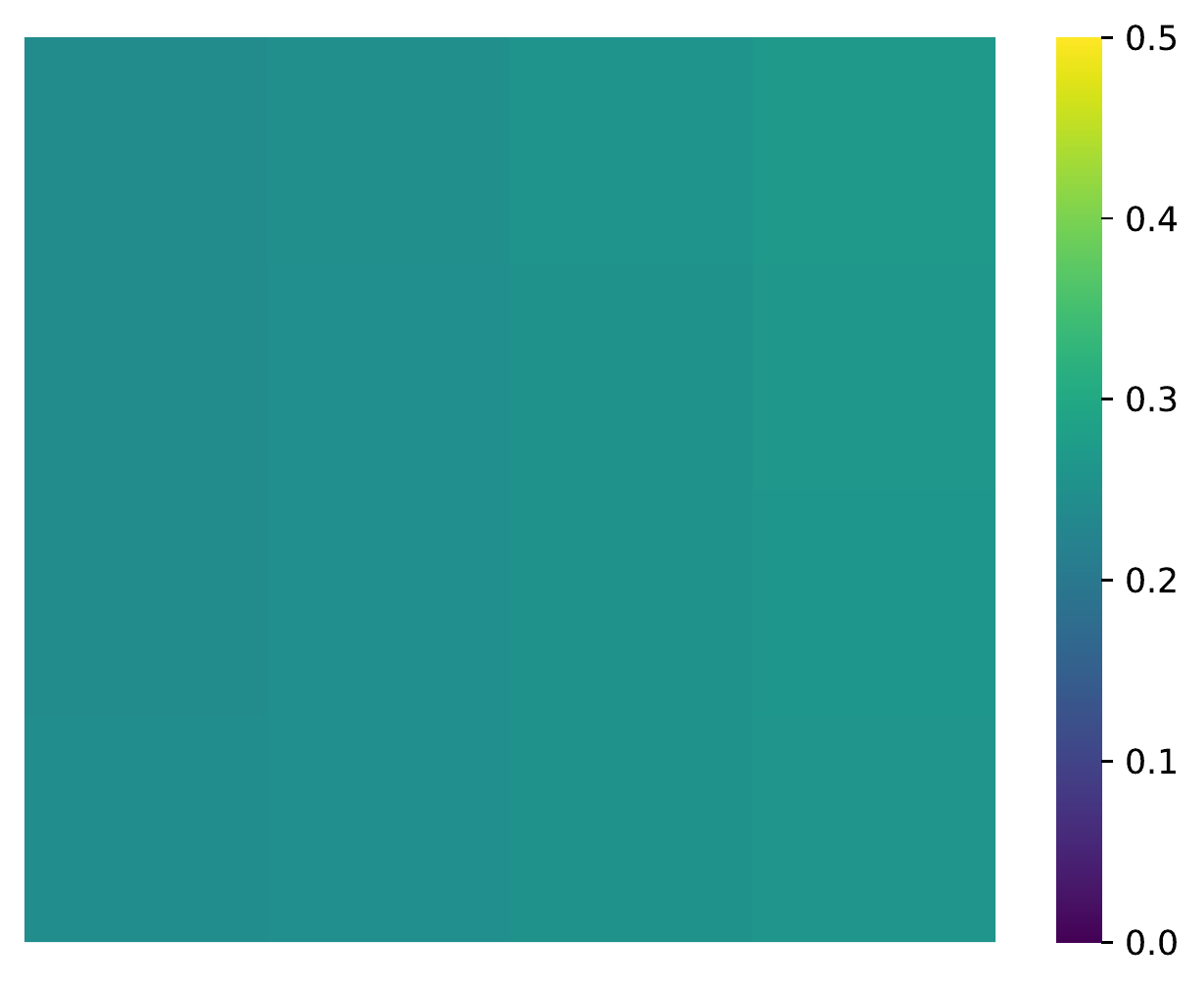}
      }\label{fig:vgg-sync-hessian}
    }
    \subfloat[{VGG-DP}]{
      {\includegraphics[scale=0.16]{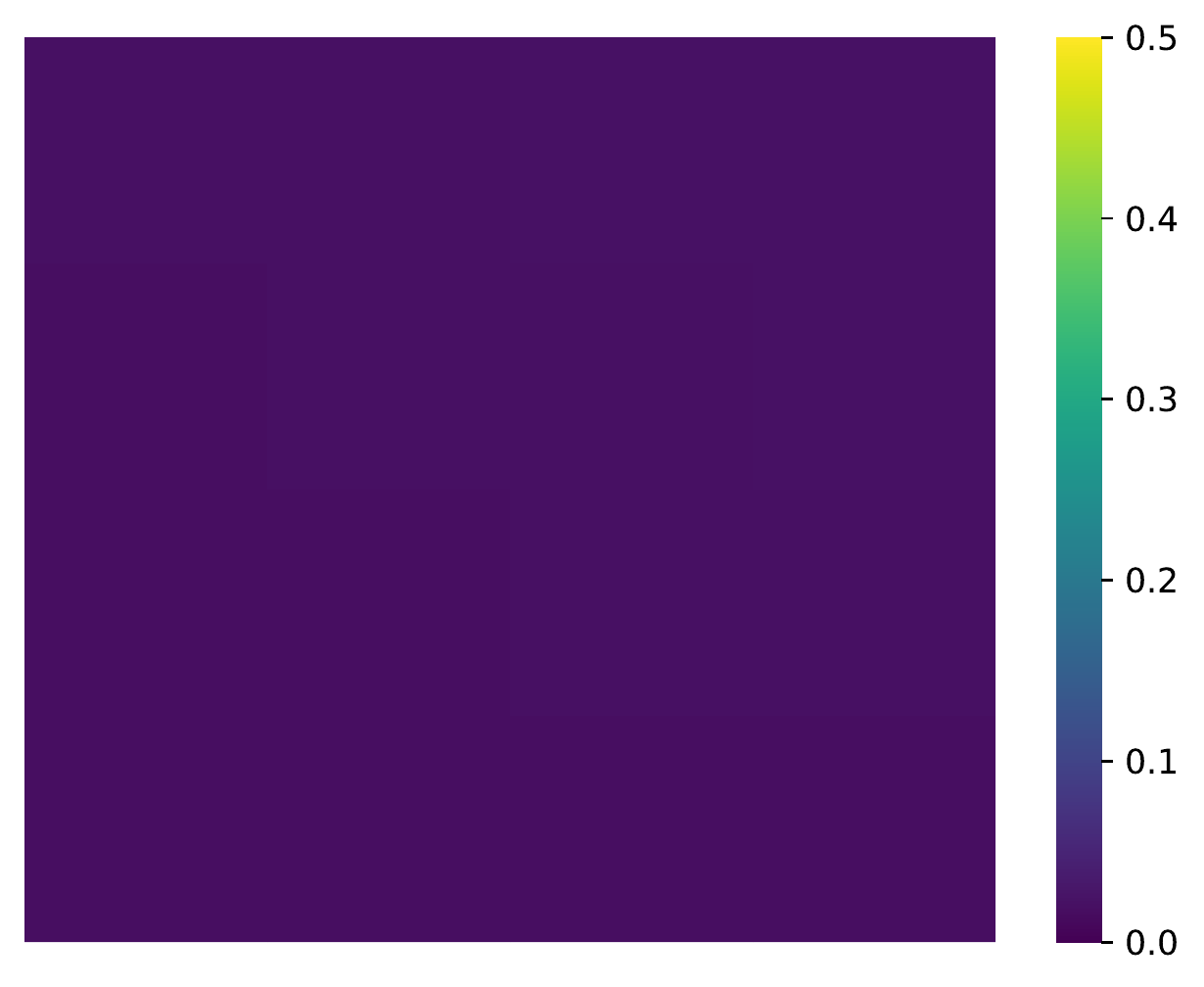}
      }\label{fig:vgg-dpsgd-hessian}
    }
    \subfloat[{ResN-S}]{
      {\includegraphics[scale=0.16]{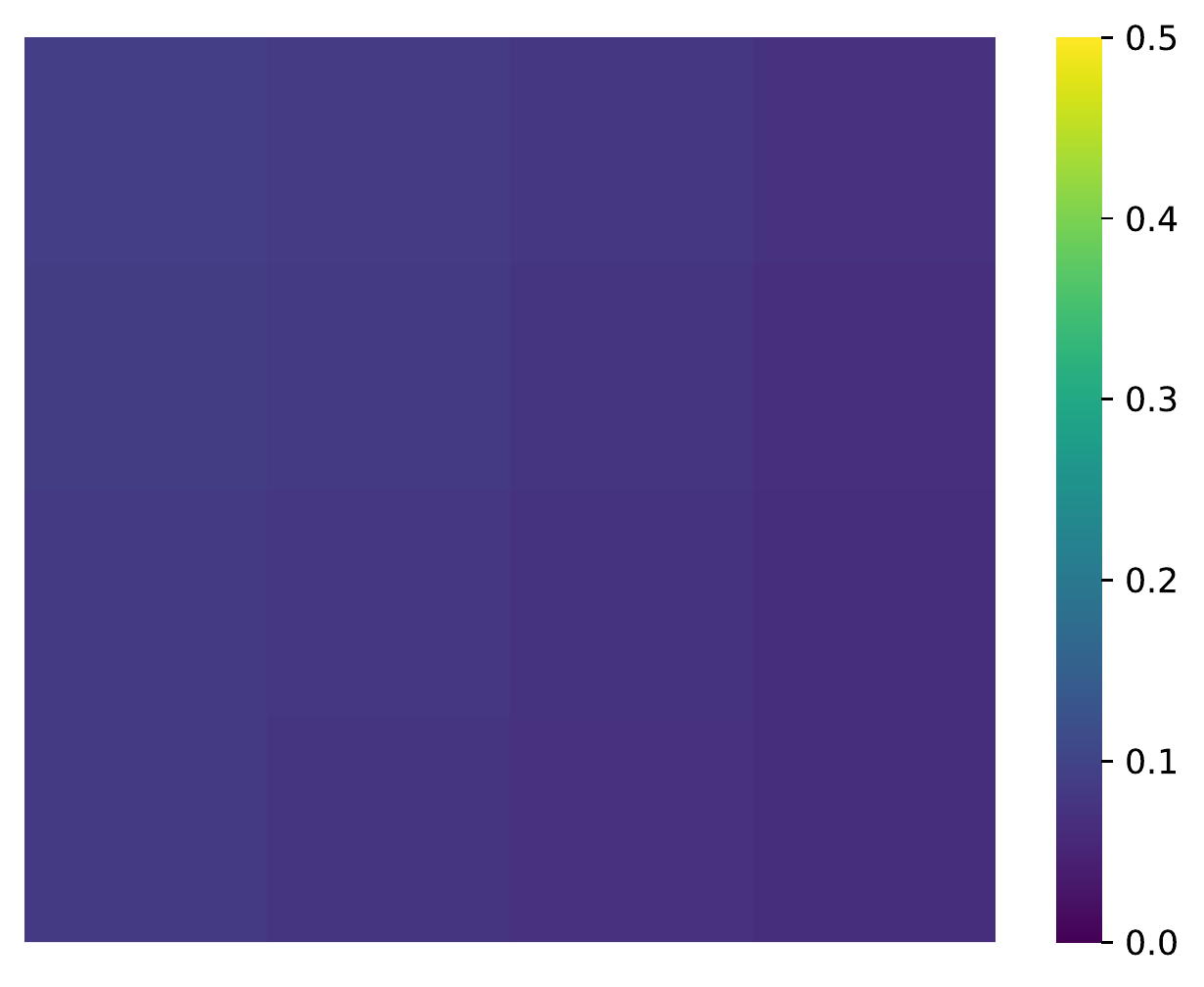}
      }\label{fig:resnet-sync-hessian}
    }
    \subfloat[{ResN-DP}]{
      {\includegraphics[scale=0.16]{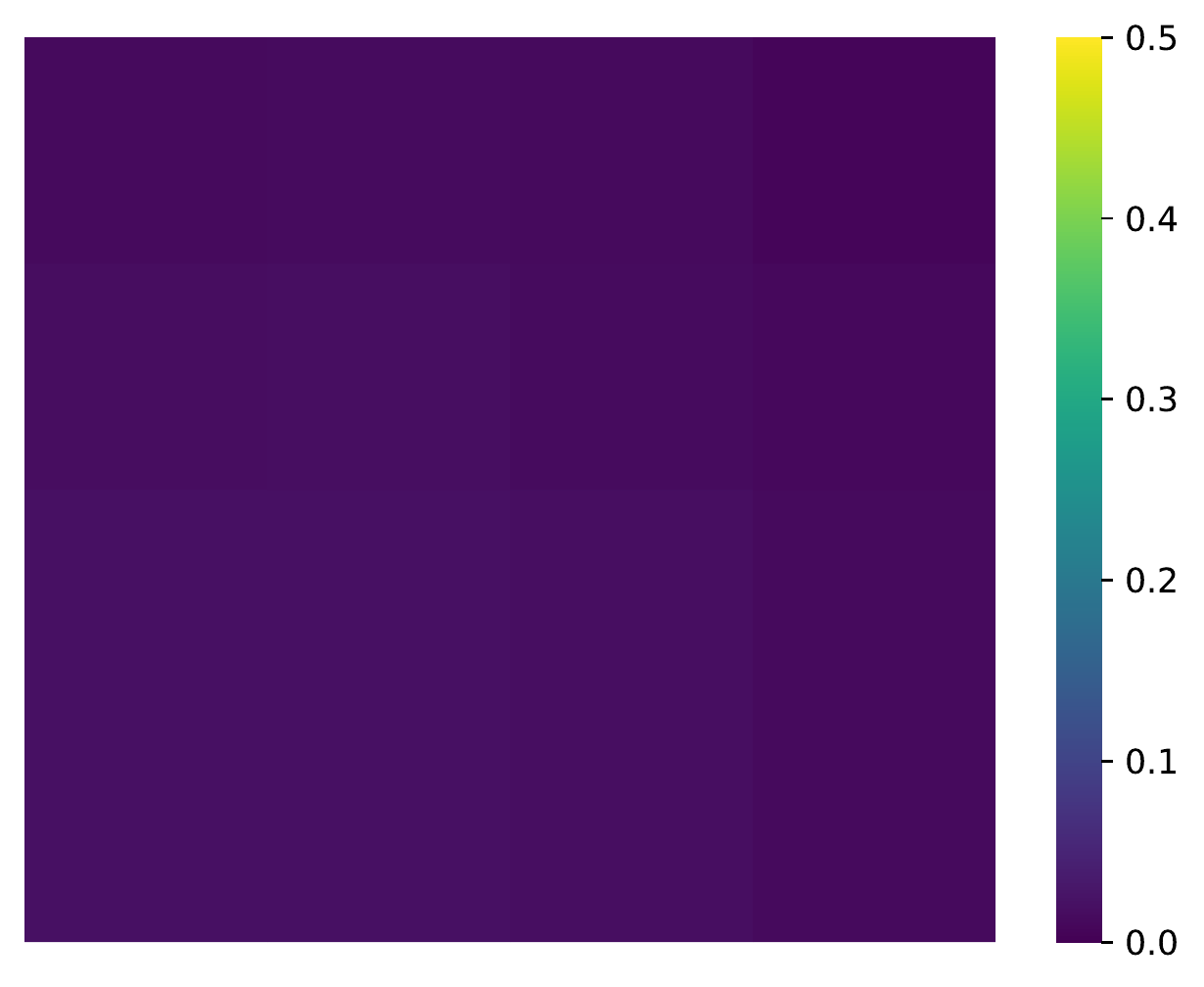}
      }\label{fig:resnet-dpsgd-hessian}
    }
    \subfloat[{Dense-S}]{
      {\includegraphics[scale=0.16]{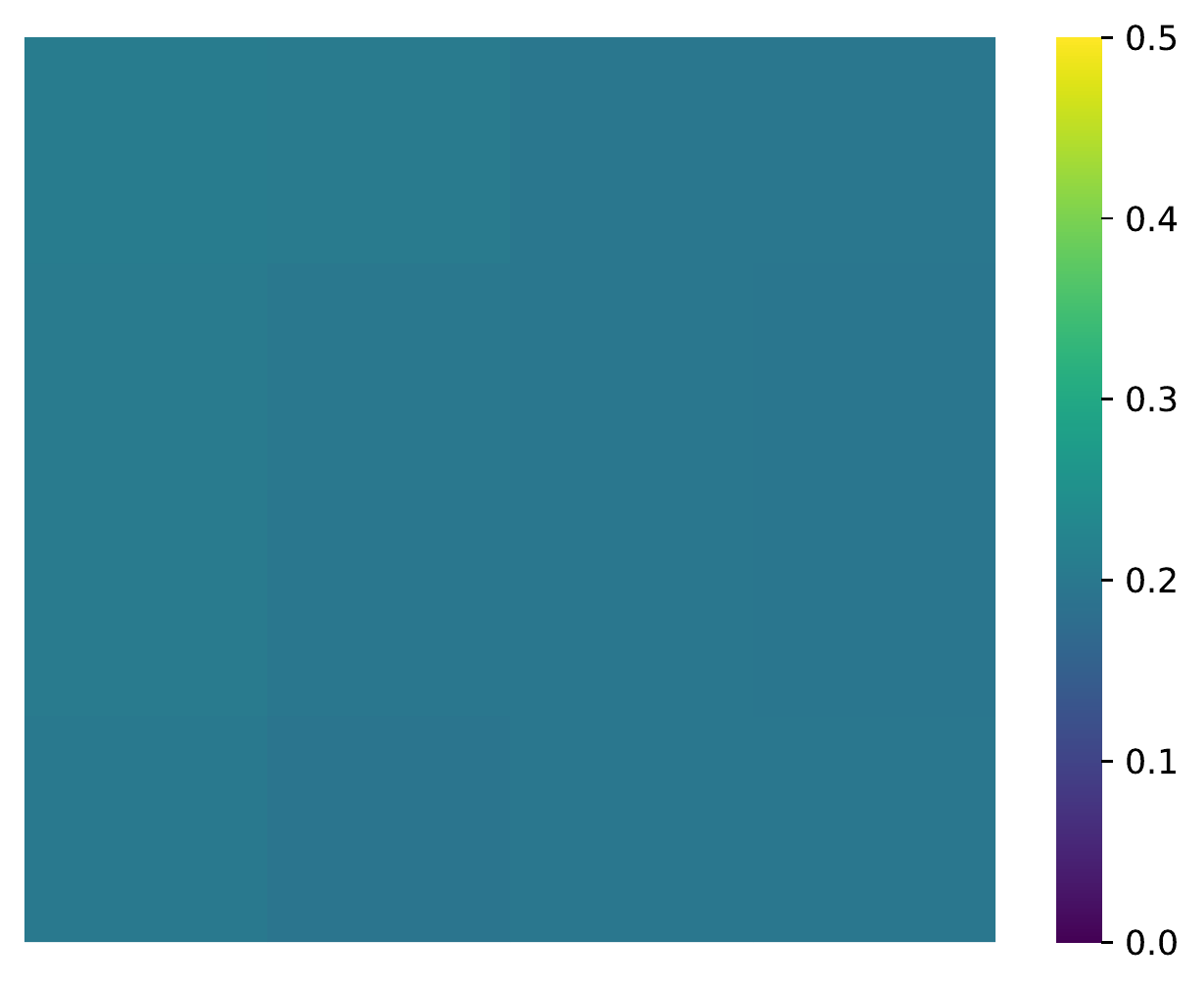}
      }\label{fig:densenet-sync-hessian}
    }
    \subfloat[{Dense-DP}]{
      {\includegraphics[scale=0.16]{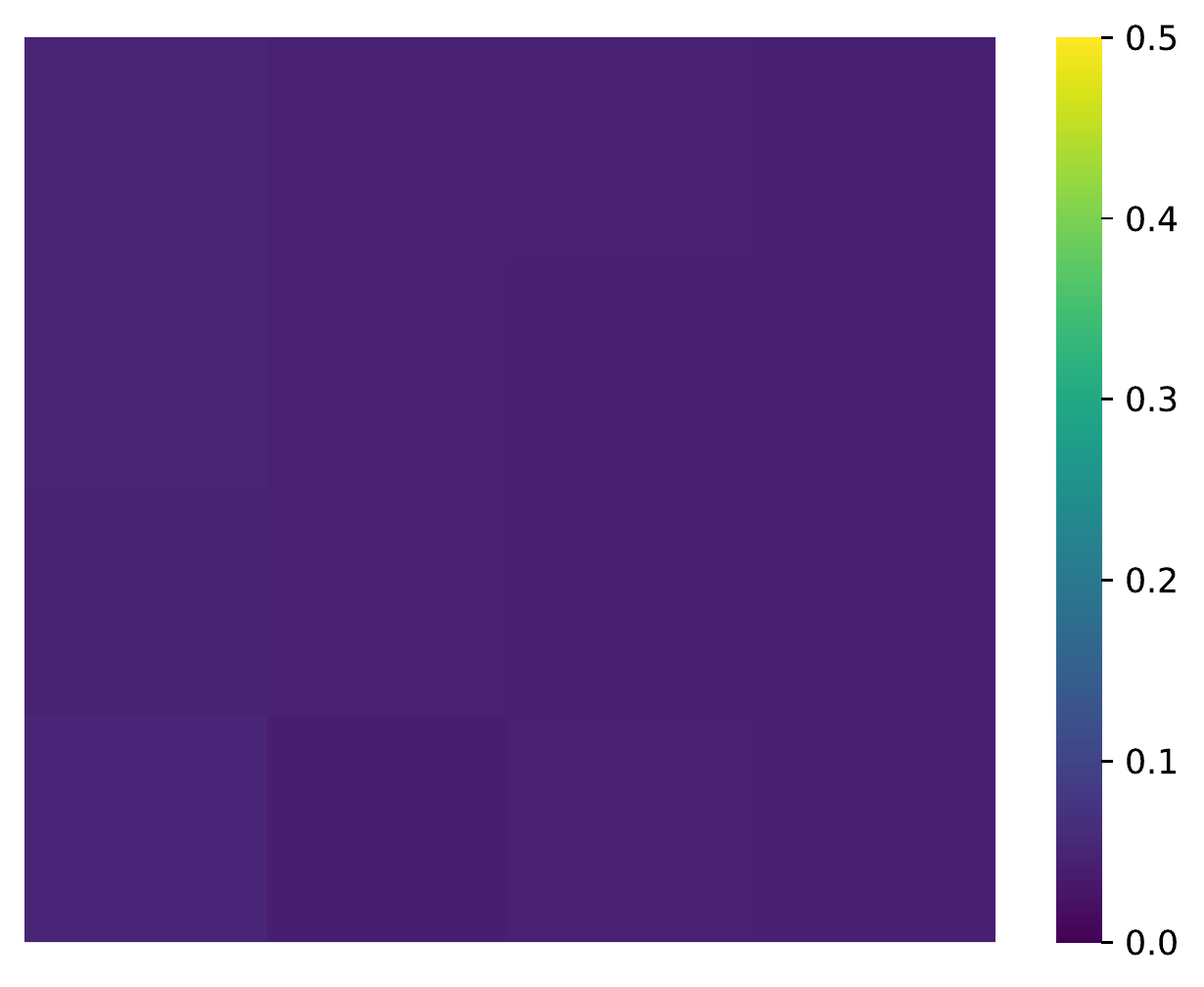}
      }\label{fig:densenet-dpsgd-hessian}
     }
    \\
\caption{\cifar Hessian heatmap on a 4x4 grid. The lower value (i.e. a cooler color) indicates the corresponding point is less likely in a saddle. We plotted against the models at the end of the 16th epoch. DPSGD is much more effective at avoiding early traps (e.g., saddle points) than SSGD.  VGG: \vgg, ResN: \resnet, DenseN: \densenet, -S: -\sync, -DP: -\dpsgd}
    \label{fig:cifar_hessian}
\end{figure*}

To better understand the loss landscape in \sync and \dpsgd training, we visualize the landscape contour 2D projection and Hessian 2D projection, using the same mechanism as in \citep{li-loss-nips}. For both plots, we randomly select two  $N$-dim vectors (where $N$ is the number of parameters in each model) and multiply with a scaling factor evenly sampled from -0.1 to 0.1 in a $K\times K$ grid to generate $K^2$ perturbations of the trained model. To produce a contour plot, we calculate the testing data loss of the perturbed model at each point in the $K\times K$ grid. \Cref{fig:cifar_2dcontour} depicts the 2D contour plot for representative models (at the end of the 320th epoch) in a $50\times 50$ grid. \dpsgd training leads not only to a lower loss but also much more widely spaced contours, indicating a flatter loss landscape and more generalizable solution. For the Hessian plot, we first calculate the maximum eigen value $\lambda_{\text{max}}$ and minimum eigen value $\lambda_{\text{min}}$ of the model's Hessian matrix at each sample point in a 4x4 grid. We then calculate the ratio $r$ between $|\lambda_{\text{min}}|$ and $|\lambda_{\text{max}}|$. The lower $r$ is, the more likely it is in a convex region and less likely in a saddle region. We then plot the heatmap of this $r$ value in this 4x4 grid. The corresponding models are trained at the 16-th epoch (i.e. the first 5\% training phase) and the corresponding Hessian plot  \Cref{fig:cifar_hessian} indicates \dpsgd is much more effective at avoiding early traps (e.g., saddle points) than \sync.

\subsection{\imagenet Training Progression}
\label{appendix-imagenet-progression}
\begin{table*}[t]
\centering
\small
\begin{tabular}{llllllll}
                &          & AlexNet   & VGG & VGG-BN   & ResNet-50  & ResNext-50 & DenseNet-161  \\
                \hline
bs=256   & Baseline & 56.31/79.05 &  69.02/88.66 & 70.65/89.92 & 76.39/93.05  & 77.62/93.64   & 78.43/94.20     \\
lr=1x  &          &  lr=0.01      &        &lr=0.1   &       &        &              \\
                \hline

bs=2048  & SSGD     & \bf{54.29/77.43} & \bf{67.67/87.91} & \bf{70.36/89.58} & \bf{76.648/92.99} & \bf{77.486/93.62}  &  \bf{78.19/94.16}     \\
lr=8x                & DPSGD    & 53.71/76.91 & 67.28/87.58 & 69.76/89.31 & 76.094/92.82 & 77.236/93.60  & 77.28/93.64     \\
\hline
bs=4096& SSGD     & 0.10/0.50 & 0.10/0.50 & 65.39/86.51 & \bf{76.46/93.06} & \bf{77.43/93.65}  & \bf{77.98/93.86}      \\
lr=16x                 & DPSGD    & \bf{52.53/76.01} & \bf{66.44/87.20} & \bf{68.86/88.82} & 75.784/92.82 & 77.24/93.54  & 77.73/93.81      \\
\hline
bs=8192 & SSGD     & 0.10/0.50    & 0.10/0.50    & 0.10/0.50 & \bf{76.096/92.80} & 76.564/93.16  & \bf{77.34/93.65}    \\
lr=32x                & DPSGD    & \bf{49.01/73.00} & \bf{65.00/86.11} & \bf{63.55/85.43} & 75.618/92.75 & \bf{77.162/93.42}  & 77.22/93.61   
\end{tabular}
\caption{ImageNet-1K Top-1/Top-5 model accuracy (\%) comparison for batch size 2048, 4096 and 8192. All experiments are conducted on 16 GPUs (learners), with batch size per GPU 128, 256 and 512 respectively. Bold texts represent the best model accuracy achieved given the specific batch size and learning rate. The batch size 256 baseline is presented for reference. bs stands for batch-size, lr stands for learning rate. Baseline lr is set to 0.01 for AlexNet and VGG11, 0.1 for the other models. In the large batch setting, we use learning rate warmup and linear scaling as prescribed in \citep{facebook-1hr}. For rough loss landscape like AlexNet and VGG, \ssgd diverges when batch size is large whereas \dpsgd converges.}
\label{tab:imagenet_comparison_full}
\end{table*}

\begin{figure*}
\centering
   
  \subfloat[\imagenet Top-1, bs=2048, lr=8x]{
    {\includegraphics[width=0.95\columnwidth]{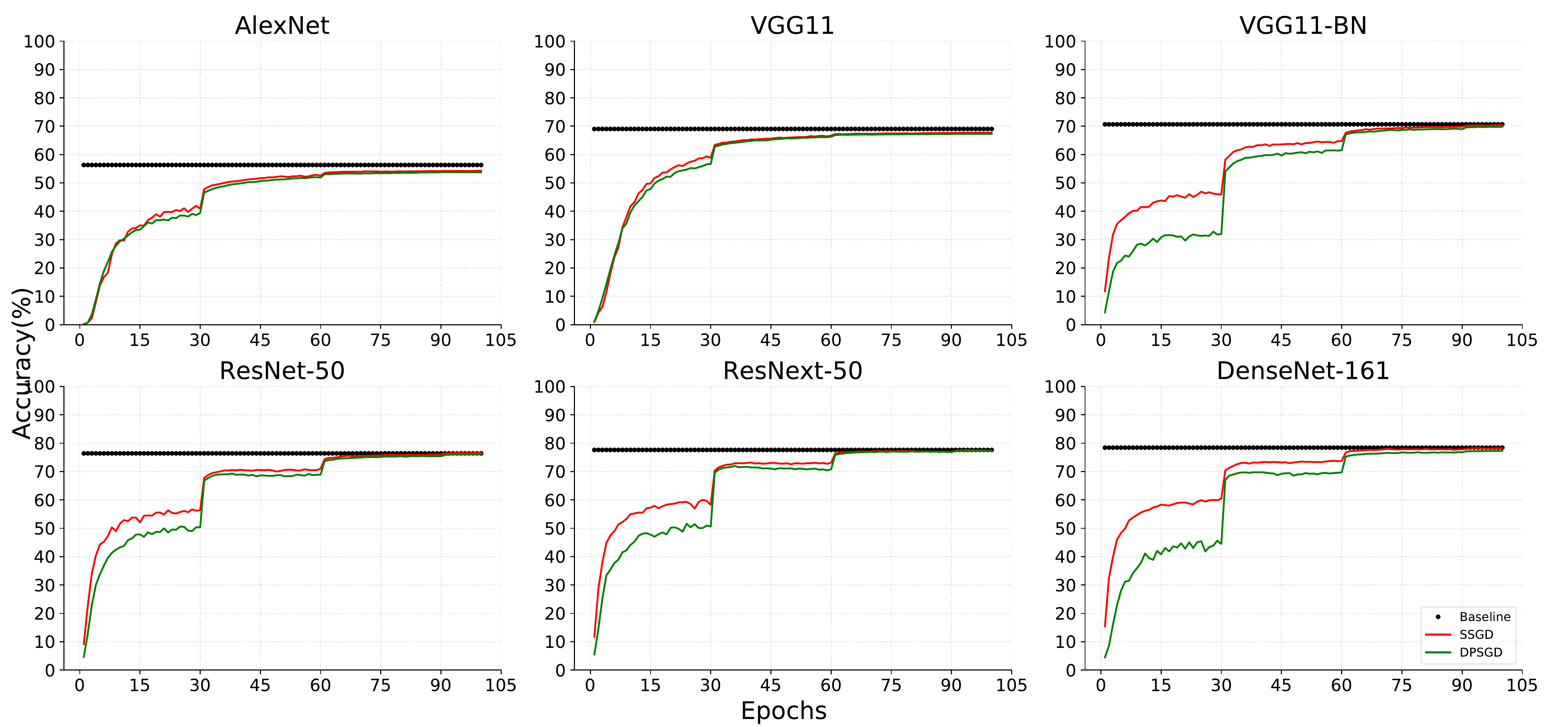}   
    }\label{fig:imagenet-convergence-bs2048}
  }
  \subfloat[\imagenet Top-5, bs=2048, lr=8x]{
    {\includegraphics[width=0.95\columnwidth]{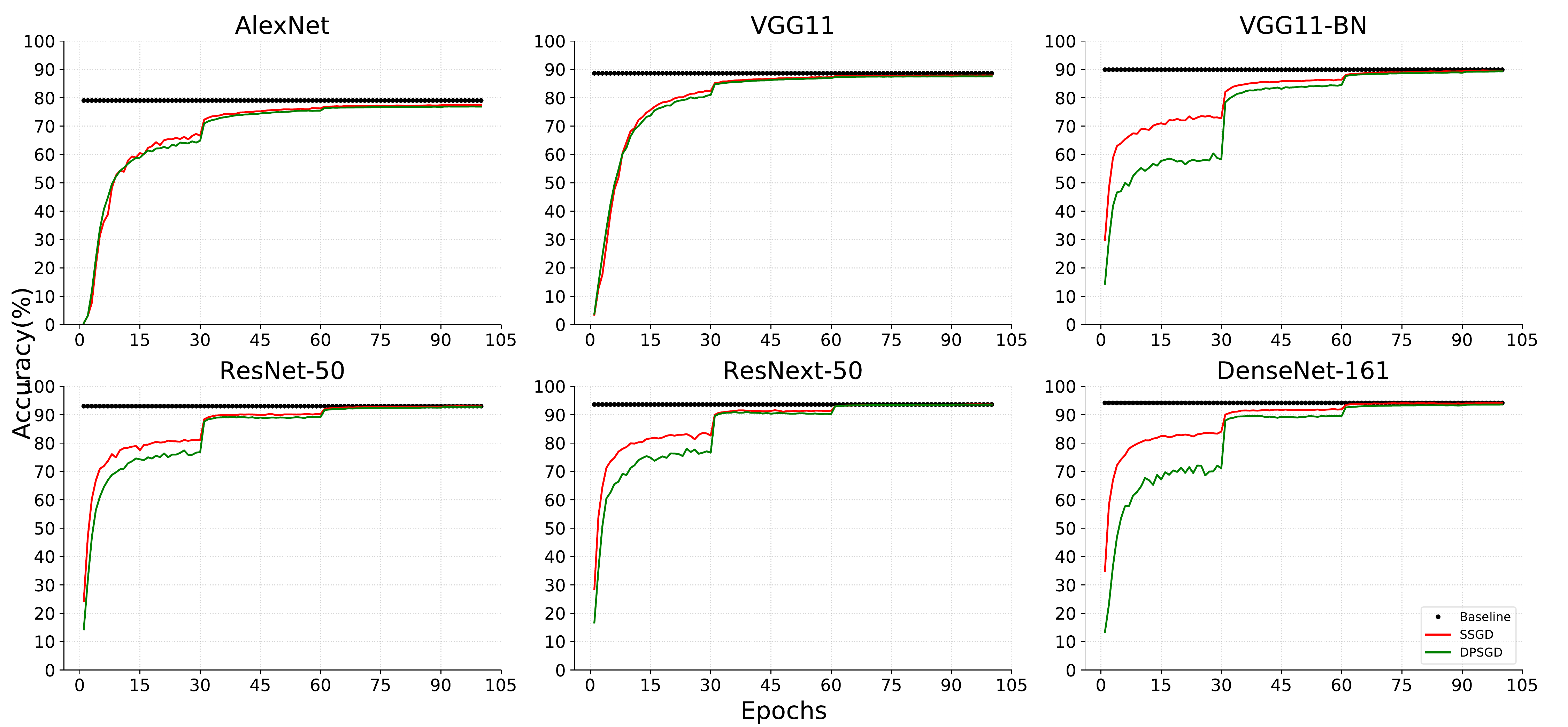}   
    }\label{fig:imagenet-convergence-bs2048-top5}
  }
  \\
  \subfloat[\imagenet Top-1, bs=4096, lr=16x]{
    {\includegraphics[width=0.95\columnwidth]{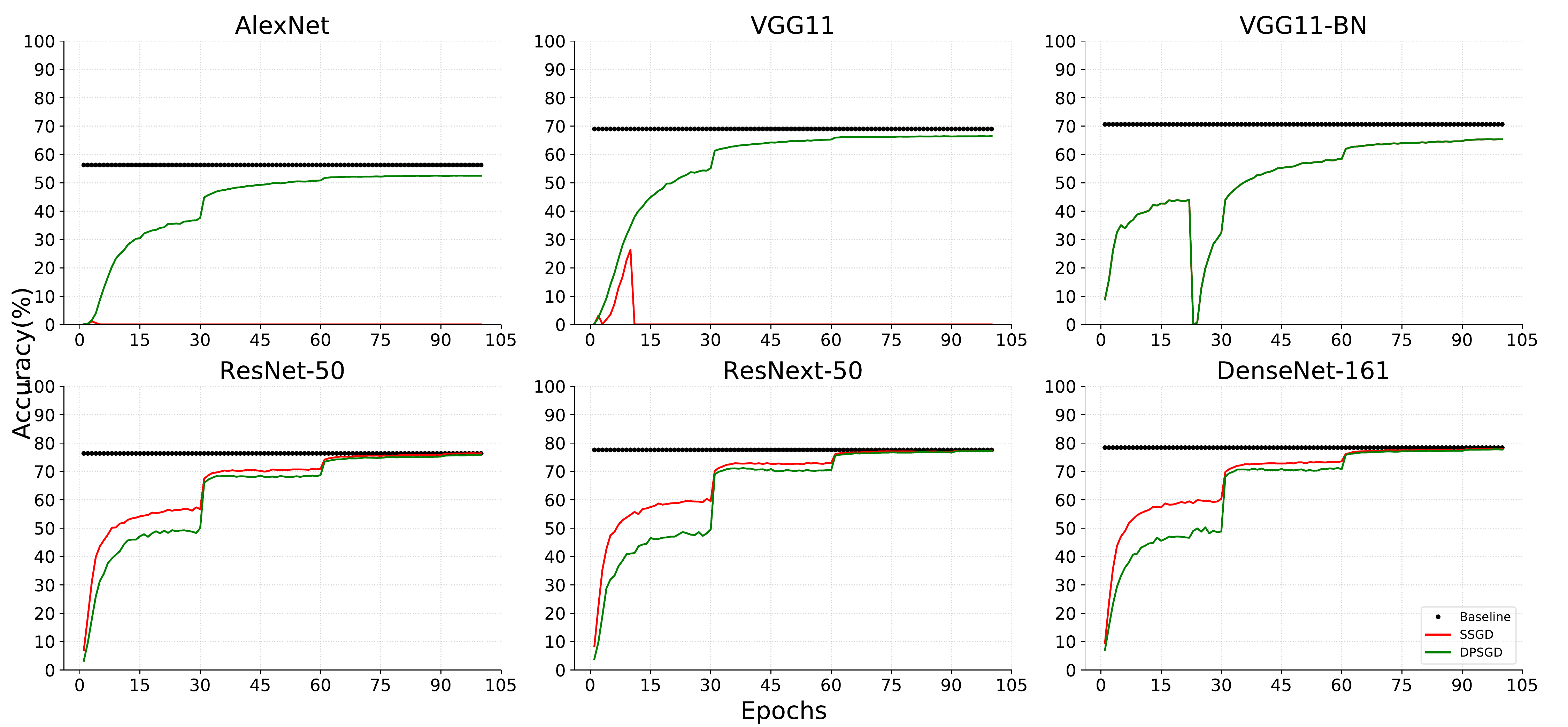}   
    }\label{fig:imagenet-convergence-bs4096}
  }
  \subfloat[\imagenet Top-5, bs=4096, lr=16x]{
    {\includegraphics[width=0.95\columnwidth]{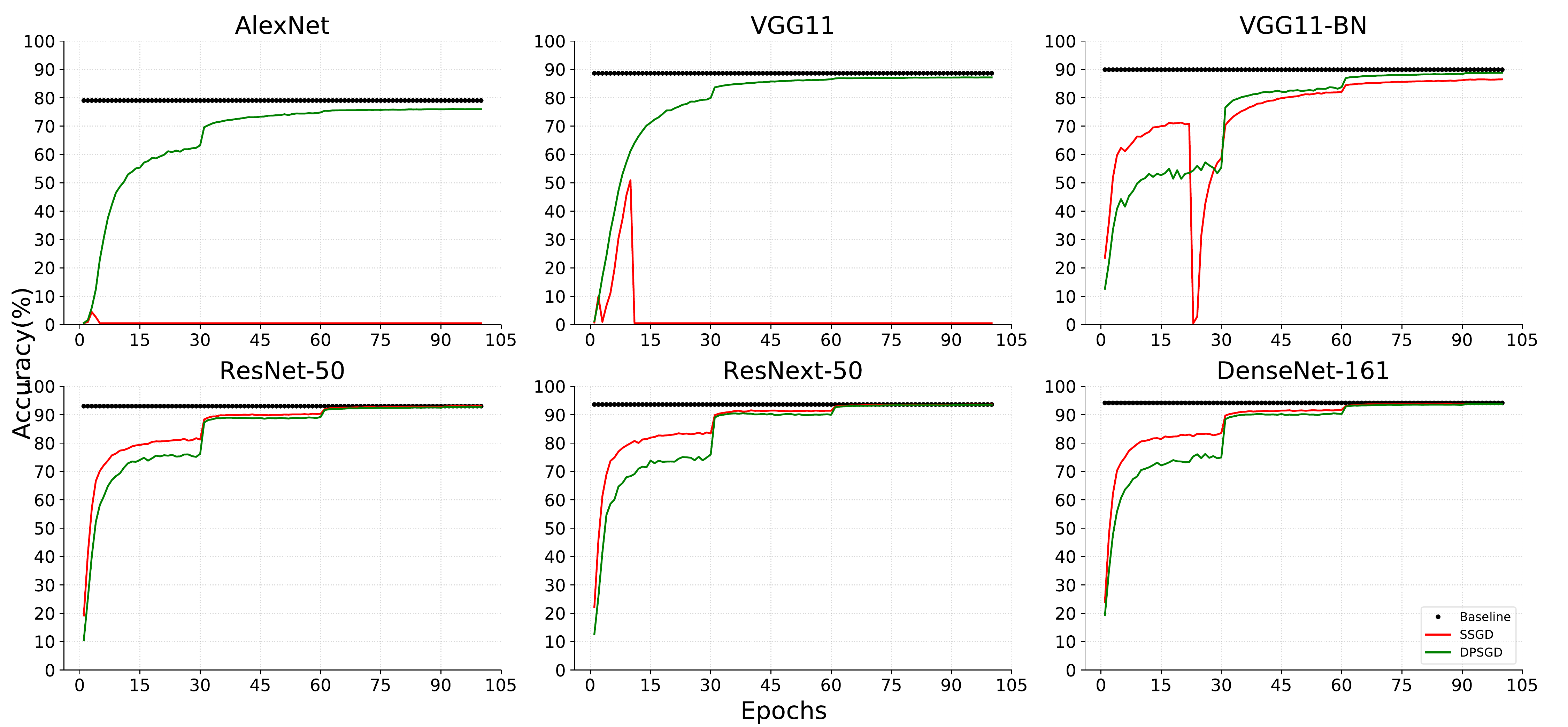}   
    }\label{fig:imagenet-convergence-bs4096-top5}
  }
  \\
  \subfloat[\imagenet Top-1, bs=8192, lr=32x]{
    {\includegraphics[width=0.95\columnwidth]{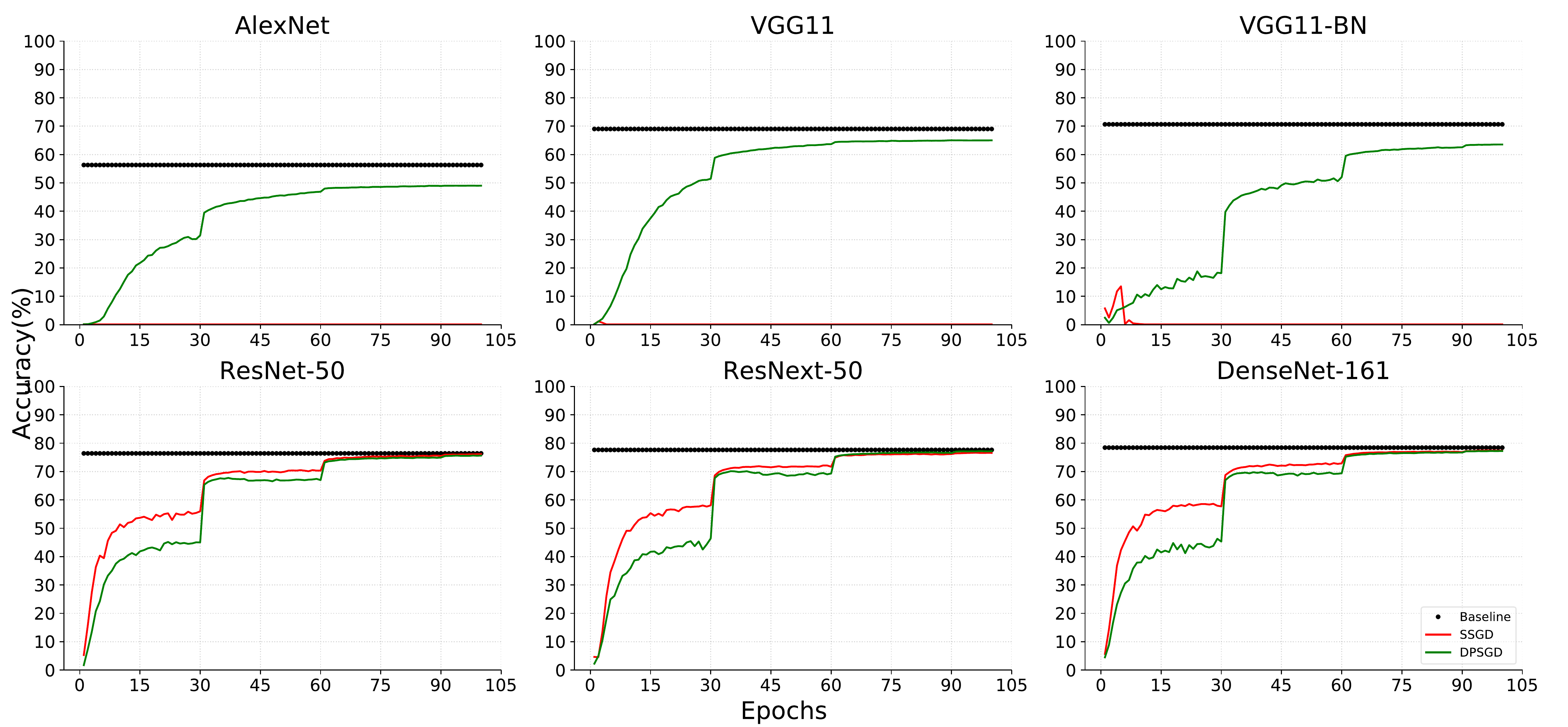}    
    }
    \label{fig:imagenet-convergence-bs8192}
  }
  \subfloat[\imagenet Top-5, bs=8192, lr=32x]{
    {\includegraphics[width=0.95\columnwidth]{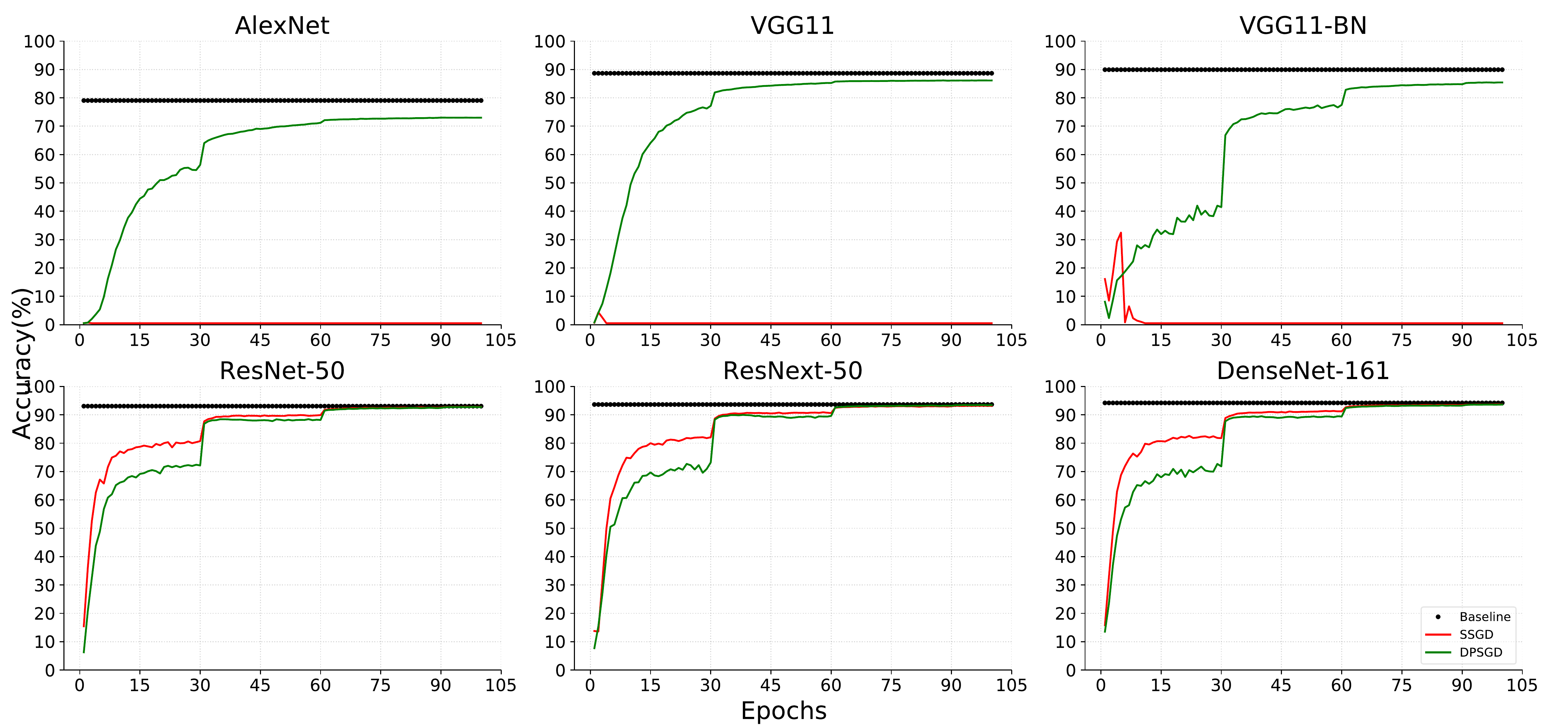}    
    }
    \label{fig:imagenet-convergence-bs8192-top5}
  }
  \caption{ImageNet-1K SSGD DPSGD comparison for batch size 2048, 4096 and 8192, with learning rate set as 0.2, 0.4 and 0.8 respectively. All experiments are conducted on 16 GPUs (learners), with batch size per GPU 128,256 and 512 respectively. When batch size is 8192, \dpsgd significantly outperforms \sync. bs stands for batch-size, lr stands for learning rate. The dotted black line represents the bs=256 baseline. }
  \label{fig:imagenet_large_bs}
\end{figure*}
\Cref{fig:imagenet_large_bs} illustrates \sync and \dpsgd comparison for \imagenet. Noticeably, \sync diverges in AlexNet, VGG11, VGG11-BN when batch-size is 8192 while \dpsgd converges.

\subsection{SWB Training Progression}
\label{appendix-swb-progression}
\begin{figure*}[t]
\small
    \centering
    \subfloat[{SWB300}]{
      {\includegraphics[width=2.0\columnwidth]{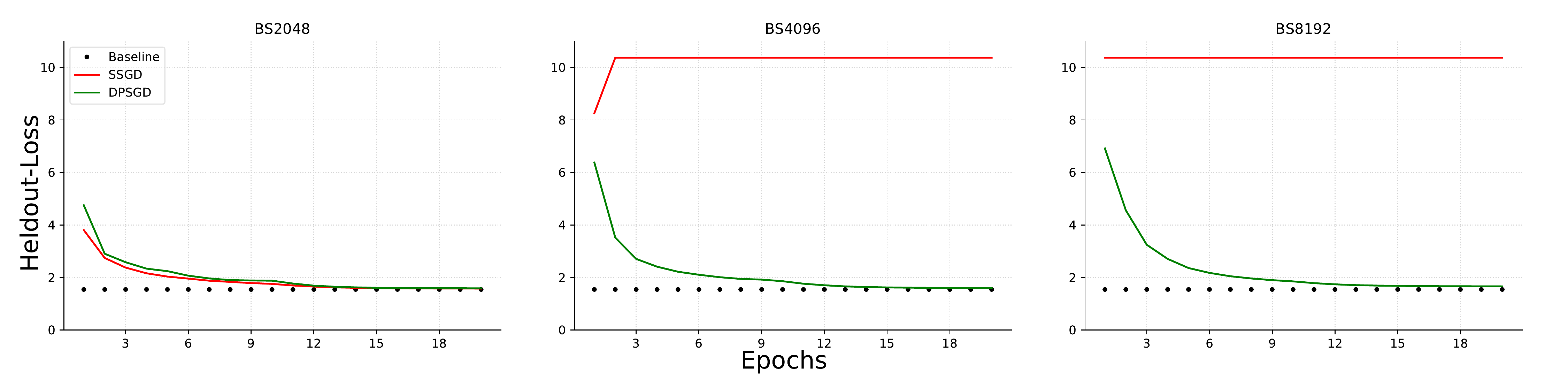}
      }\label{fig:eff-sync-e1}
    } \\
    \subfloat[{SWB2000}]{
      {\includegraphics[width=2.0\columnwidth]{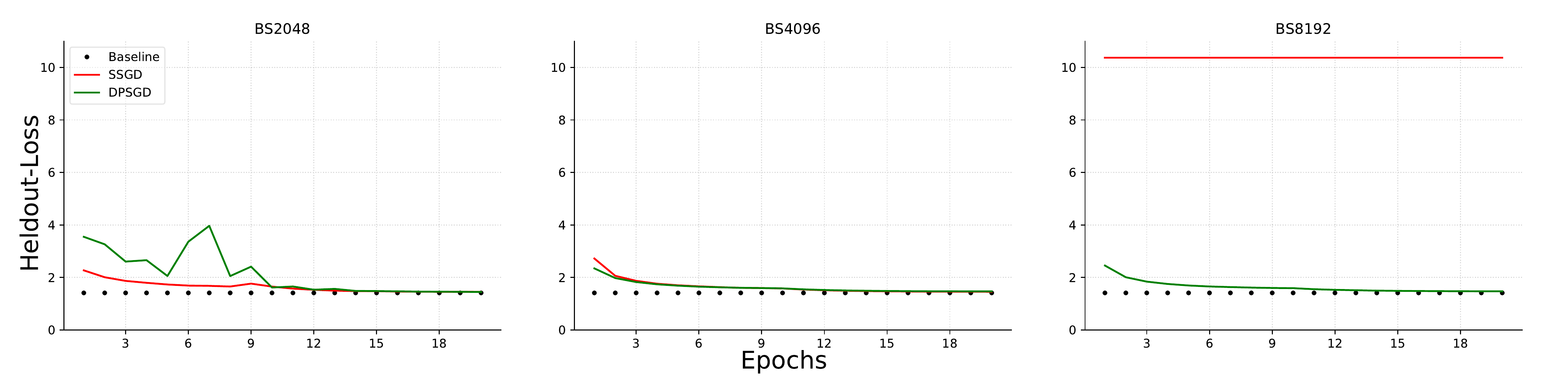}
      }\label{fig:eff-dpsgd-e1}
    }
\caption{Heldout loss w.r.t epochs for \swba and \swbb. Dotted black lines indicate the batch size 256 heldout loss baseline.}
    \label{fig:asr_loss}
\end{figure*}
\Cref{fig:asr_loss} illustrates heldout loss comparison for \swba and \swbb. In \swba task, \sync diverges beyond batch size 2048 and \dpsgd converges well til batch size 8192. In \swbb task, \sync diverges beyond batch size 4096 and \dpsgd converges well til at least batch size 8192.

\section{Appendix: End-to-End Run-time Comparison and  Advice for Practitioners}
\label{appendix:runtime}
\paragraph{End-to-End Run-time Comparison}
In all above-mentioned DPSGD and SSGD experiments we used the \textit{same} number of epochs as in the well-tuned single-GPU baseline (i.e., the total computation cost is fixed).
When computation cost is fixed, DPSGD inherently runs faster than SSGD because DPSGD requires less messages transmitted and tolerate high-latency network better ~\citep{dpsgd}. \Cref{tab:rt} records training time for each representative task (batch size 128 per GPU, 16 GPUs) on both low and high latency networks. Other tasks and batch-size setups show the same trend: DPSGD runs faster than SSGD. Further note that for Eff-B0 (target accuracy 90\%) and SWB-2000 (target heldout loss 1.48), DPSGD reaches target model quality with twice the batch size as used in SSGD, all learning rates considered (\Cref{tab:cifar10_bs8192_compare_lr} , \Cref{tab:swb_compare_lr}). Thus DPSGD can effectively use 2X more GPUs. DPSGD achieves target accuracy for Eff-B0 in 0.067 hours and for SWB-2000 in 10.08 hours (64 GPUs). In contrast, SSGD achieves target accuracy for Eff-B0 in 0.19 hours and for SWB-2000 in 23.15 hours (32 GPUs). 

 \textit{Summary} DPSGD consistently runs faster than SSGD to reach target accuracy in the large batch setting.

\begin{table*}[bhtp]
\footnotesize
\centering
\begin{tabular}{|c|c|c|c|c|c|c|c|c|c|}
\hline
                                                                    &            & Eff-b0 & SE-18 & Res-18 & Dense-121 & Mobile & Google & ResNext-29 & SWB-2000 \\ \hline
                                                                    & Single-GPU & 2.92   & 1.58  & 1.37   & 5.48      & 1.02   & 5.31   & 4.55       & 203.21   \\ \hline
\multirow{2}{*}{\begin{tabular}[c]{@{}l@{}}Latency\\ (1$\mu$s)\end{tabular}} & SSGD       & 0.34   & 0.39  & 0.35   & 0.68      & 0.17   & 0.58   & 0.56       & 38.00    \\ \cline{2-10} 
                                                                    & DPSGD      & 0.26   & 0.31  & 0.32   & 0.58      & 0.12   & 0.49   & 0.41       & 29.71    \\ \hline
\multirow{2}{*}{\begin{tabular}[c]{@{}l@{}}Latency\\ (1$m$s)\end{tabular}} & SSGD       & 0.46   & 0.85  & 0.82   & 0.96      & 0.30   & 0.84   & 0.94       & 96.31   \\ \cline{2-10} 
                                                                    & DPSGD      & 0.27   & 0.31  & 0.32   & 0.58      & 0.13   & 0.50   & 0.42       & 29.85    \\ \hline
\end{tabular}
\caption{Time (hours) to complete training  with batch size 128 per GPU and 16 GPUs in total (\cifar and \swbb).}
\label{tab:rt}
\end{table*}
\paragraph{Advice for Practitioners}
In SSGD, when total batch size is fixed, the convergence behavior is the same regardless of the number of learners. In DPSGD, when the number of learners increases, the convergence could be harmed due to too much discrepancy between learners. In another word, we would like a system that has enough system noise so that it can help avoid early training traps but not too much noise so that model convergence is unaffected. In practice, we found that 16-learner setup usually yields the best convergence results in the DPSGD setting, which is consistent with research literature ~\citep{dpsgd, adpsgd}. To make use of a larger number of computing devices in DPSGD, we recommend a hierarchical system design ~\citep{interspeech19} where we group nearby learners (e.g., on the same server) as one big super-learner and apply DPSGD algorithm only across super-learners. For example, on a 128 GPU cluster, we could group 8 learners as one big super-learner and we apply DPSGD among 16 super-learners. In addition, we also recommend in each iteration, each (super)-learner selects a random neighbor to communicate to further improve convergence. Please refer to ~\citep{icassp20} for the detailed analysis of how randomized communication  improves DPSGD convergence.

\end{document}